\PassOptionsToPackage{table}{xcolor}
\documentclass{article} 
\usepackage{iclr2026_conference,times}

\usepackage{amsmath}
\usepackage{amssymb}
\usepackage{mathtools}
\usepackage{amsthm}
\usepackage{wasysym}
\usepackage{microtype}
\usepackage{graphicx}
\usepackage{subcaption}
\usepackage{booktabs} 
\usepackage{multirow}
\usepackage{wrapfig}
\usepackage{enumitem}

\usepackage[utf8]{inputenc}
\usepackage{xcolor}
\usepackage{array}
\usepackage[table]{xcolor}

\usepackage{hyperref}

\definecolor{growthgreen}{rgb}{0.0, 0.5, 0.0}

\usepackage[capitalize,noabbrev]{cleveref}

\theoremstyle{plain}

\theoremstyle{definition}

\theoremstyle{remark}

\makeatletter
\DeclareRobustCommand{\iscircle}{\mathord{\mathpalette\is@circle\relax}}
\newcommand\is@circle[2]{%
  \begingroup
  \sbox\z@{\raisebox{\depth}{$\m@th#1\bigcirc$}}%
  \sbox\tw@{$#1\square$}%
  \resizebox{!}{\ht\tw@}{\usebox{\z@}}%
  \endgroup
}
\makeatother

\makeatletter
\DeclareRobustCommand{\ishexagon}{\mathord{\mathpalette\is@hexagon\relax}}
\newcommand\is@hexagon[2]{%
  \begingroup
  \sbox\z@{\raisebox{\depth}{$\m@th#1\hexagon$}}%
  \sbox\tw@{$#1\square$}%
  \resizebox{!}{\ht\tw@}{\usebox{\z@}}%
  \endgroup
}
\makeatother

\makeatletter
\DeclareRobustCommand{\istriangle}{\mathord{\mathpalette\is@triangle\relax}}
\newcommand\is@triangle[2]{%
  \begingroup
  \sbox\z@{\raisebox{\depth}{$\m@th#1\triangle$}}%
  \sbox\tw@{$#1\square$}%
  \resizebox{!}{\ht\tw@}{\usebox{\z@}}%
  \endgroup
}

\DeclareRobustCommand{\issquare}{\mathord{\mathpalette\is@square\relax}}
\newcommand\is@square[2]{%
  \begingroup
  \sbox\z@{\raisebox{\depth}{$\m@th#1\square$}}%
  \sbox\tw@{$#1\square$}%
  \resizebox{!}{\ht\tw@}{\usebox{\z@}}%
  \endgroup
}
\makeatother

\usepackage{xspace}         

\newcommand{\model}{\textsc{Ear}\xspace} %
\newcommand{\modelfull}{Editing as Reasoning\xspace} %
\newcommand{\bench}{\textsc{Amaze}\xspace} %

\usepackage{tcolorbox}

\newtcolorbox{definitionbox}{
  colback=blue!5!white,  
  colframe=blue!75!black, 
  sharp corners,         
  boxrule=1pt,           
}

\newtcolorbox{definitionbox2}{
  colback=green!5!white,  
  colframe=green!25!black, 
  sharp corners,         
  boxrule=1pt,           
}

\usepackage{amsmath,amsfonts,bm}

\def\eqref#1{equation~\ref{#1}}

\def\1{\bm{1}}

\DeclareMathAlphabet{\mathsfit}{\encodingdefault}{\sfdefault}{m}{sl}
\SetMathAlphabet{\mathsfit}{bold}{\encodingdefault}{\sfdefault}{bx}{n}

\usepackage{threeparttable}
\usepackage{url}
\hypersetup{hidelinks} 

\usepackage[misc]{ifsym} 
\usepackage{fontawesome5}

\newcommand{\blfootnote}[1]{%
  \begingroup
  \renewcommand\thefootnote{}\footnote{#1}%
  \addtocounter{footnote}{-1}%
  \endgroup
}

\title{Probing Visual Planning\\ in Image Editing Models}

\author{Zhimu Zhou$^1$\quad Yanpeng Zhao$^{3,*}\,^\dagger$\quad Qiuyu Liao$^2$\quad Bo Zhao$^{1,*}$ \quad Xiaojian Ma$^3$\\[0.2em]
$^1$School of AI, Shanghai Jiao Tong University \quad $^2$Renmin University of China\\  $^3$State Key Laboratory of General Artificial Intelligence, BIGAI
\\[0.25em]
{\small\texttt{\{sjtu2717992; bo.zhao\}@sjtu.edu.cn, yannzhao.ed@gmail.com}}
\\[0.25em]
\faGlobe~\url{https://spatigen.github.io/amaze.io/}
\hfill
\faGithub~\url{https://github.com/spatigen/amaze}
}

\iclrfinalcopy 
\begin{document}

\maketitle

\blfootnote{
\begin{tabular}{@{}l@{}}
$\dagger$: Project Lead. $*$: Corresponding author: Bo Zhao. \\
\end{tabular}
}

%
\begin{abstract}

Visual planning represents a crucial facet of human intelligence, especially in tasks that require complex spatial reasoning and navigation. Yet, in machine learning, this inherently visual problem is often tackled through a verbal-centric lens. While recent research demonstrates the promise of fully visual approaches, they suffer from significant computational inefficiency due to the step-by-step planning-by-generation paradigm.
In this work, we present \model, an editing-as-reasoning paradigm that reformulates visual planning as a single-step image transformation. To isolate intrinsic reasoning from visual recognition, we employ abstract puzzles as probing tasks and introduce \bench, a procedurally generated dataset that features the classical Maze and Queen problems, covering distinct, complementary forms of visual planning. The abstract nature of \bench also facilitates automatic evaluation of autoregressive and diffusion-based models in terms of both pixel-wise fidelity and logical validity. We assess leading proprietary and open-source editing models. The results show that they all struggle in the zero-shot setting, finetuning on basic scales enables remarkable generalization to larger in-domain scales and out-of-domain scales and geometries. However, our best model that runs on high-end hardware fails to match the zero-shot efficiency of human solvers, highlighting a persistent gap in neural visual reasoning.

\end{abstract}

%

\vspace{-20pt} 
\begin{wrapfigure}{r}{0.45\linewidth}
\vspace{-25pt} 
    \centering
    \includegraphics[width=0.90\linewidth]{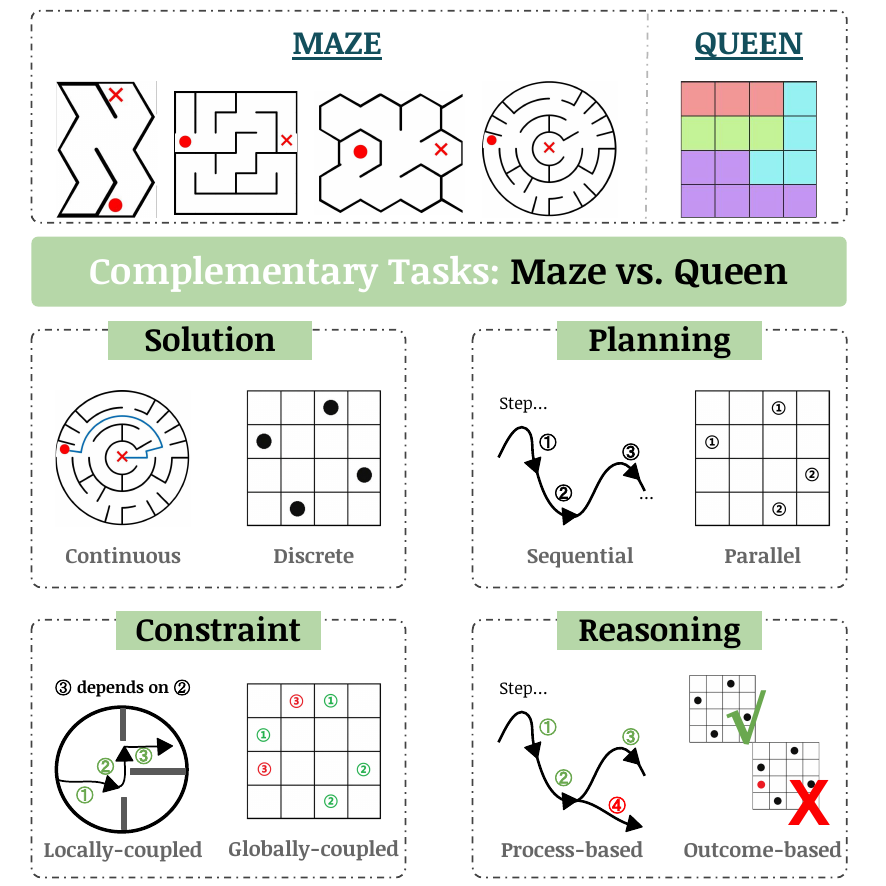} 
    \vspace{-5pt}
    \caption{The \bench tasks.}
    \label{fig:simple_pipe}
\end{wrapfigure}
\section{Introduction}

Spatial reasoning through visual planning is a cornerstone in human intelligence. While humans can navigate complex visual environments intuitively, machine learning models have been predominantly relying on verbal-centric approaches, such as translating these inherently visual reasoning problems into text for large language models (LLMs)~\citep{yang2022empirical, DBLP:journals/corr/abs-2303-04671, wang-etal-2025-mathcoder,dao2025alphamaze} and framing them as multimodal tasks that rely on vision-language models for text-based chain-of-thought~\citep{pmlr-v202-li23q, Xu_2025_ICCV, zhang2025reasongen, zhang-etal-2025-improve,wu2025reinforcing}. Recently, reasoning-enhanced generative image models have enabled fully visual alternatives. Some approaches utilize step-wise image-level generation to implement planning but suffer from significant computational inefficiency~\citep{xu2025visual}; while others attempt direct-generation methods~\citep{videozeroshot}, yet a comprehensive understanding of the intrinsic visual planning capabilities within these editing-based models remains elusive.  

\begin{figure*}[t!]
    \vspace{-30pt} 
    \centering
    \includegraphics[width=0.9\textwidth]{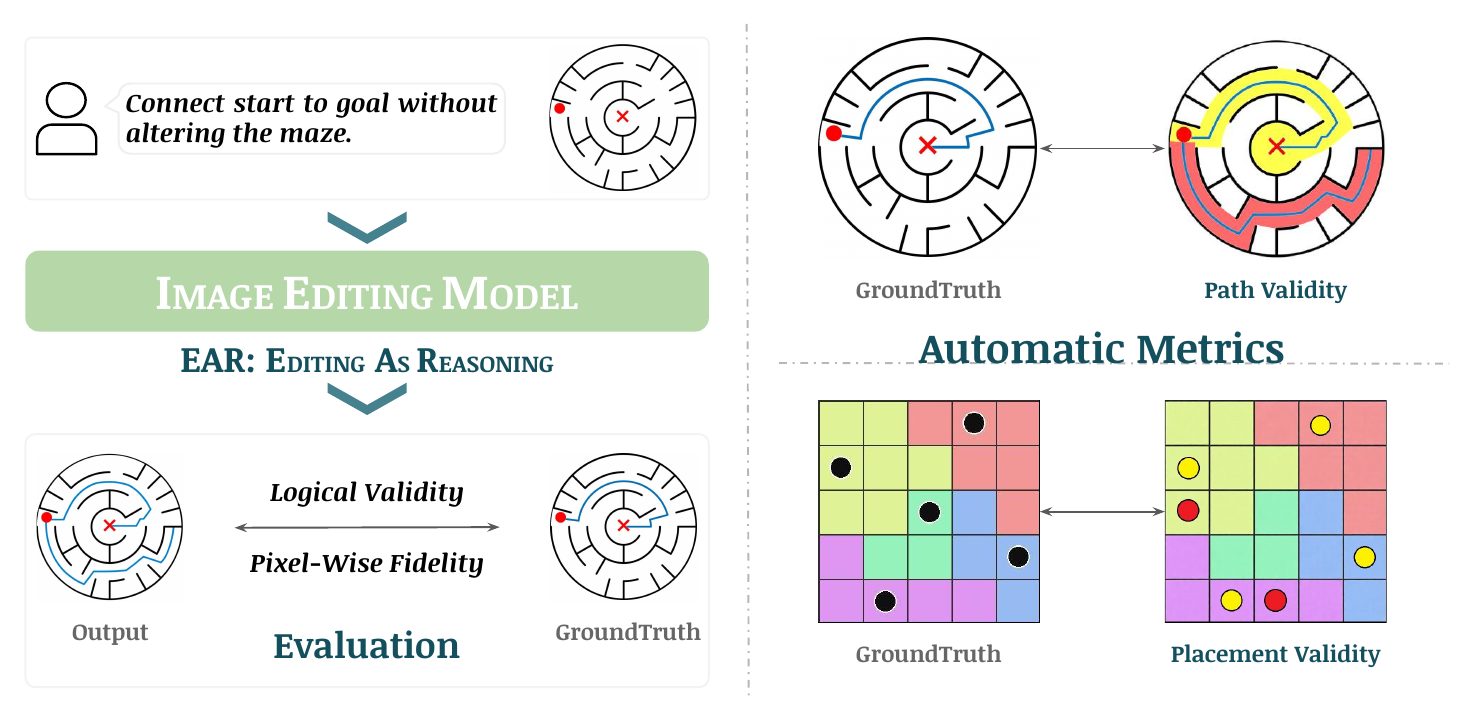} 
    \caption{Overview of \model. Left: the \model paradigm. Right: automatic evaluation. Yellow and red highlight the generated image's overlap with the solution and non-solution areas, respectively.
    }
    \label{fig:metric_example} 
    \vspace{-12pt} 
\end{figure*}

To bridge this gap, we present \modelfull (\model), a fully visual reasoning framework that reformulates visual planning as an image editing task. Unlike step-wise approaches, \model compresses the planning process into an atomic ``edit", leveraging the model's internalized spatial and visual priors to produce a complete solution in a single step. By offloading planning to the inherent progressive dynamics of the atomic ``edit", \model eliminates the inductive bias of explicit step-wise modeling, enabling a targeted probing of the intrinsic visual planning capabilities of editing models.

To facilitate in-depth and controlled analysis, we introduce \bench, a procedurally generated benchmark for visual planning. \bench comprises Maze and Queen tasks that respectively covers two complementary planning paradigms: sequential planning under local constraints and combinatorial planning under global constraints (see Figure~\ref{fig:simple_pipe}). \bench isolates intrinsic visual reasoning from the confounding factor of complex visual recognition.
Its abstract nature enables automatic evaluation metrics that decouple visual reconstruction (pixel-wise fidelity) from logical validity (topological correctness or constraint satisfaction).
We incorporate Queen puzzles across different scales (e.g., $4\times4$ and $10\times10$), with Maze further featuring diverse geometry types (including triangle, square, hexagon, and circle) to represent varying levels of complexity. This structural diversity enables us to probe the geometric invariance and systematicity of neural visual reasoning, assessing whether models develop generalizable spatial logic or merely exploit local patterns.

We evaluate representative autoregressive and diffusion-based editing models from both proprietary and open-source domains. Our probing experiments are organized along three primary dimensions: (1) \emph{Generalizability} evaluates how well models transfer to unseen geometry types and scales, including both in-domain and out-of-domain settings. (2) \emph{Scaling effect} investigates the scaling law in fine-tuned models for enhanced visual planning, i.e., relationships between performance and quantities of training data and time. (3) \emph{Human comparison} benchmarks the efficiency of visual planning of editing models against human solvers to reveal the performance gap.

Our evaluation reveals that both proprietary and open-source editing models initially struggle with zero-shot visual planning. On Maze, while proprietary models are relatively stronger, finetuning open-source models like Bagel~\citep{deng2025bagel} on basic $3\times3$ mazes improves from 0 to 11.54\% (\textsc{Pass@1}), outperforming the best proprietary model by an \emph{absolute} 6.14\%, and the fine-tuned models show impressive generalizability to larger scales. Notably, diffusion-based models surpass autoregressive models on both Maze and Queen after fine-tuning, suggesting their effectiveness in developing visual reasoning logic. 
Moreover, our comparison with human solvers on \bench reveals a stark efficiency gap: our best model, when running on a single NVIDIA RTX 5090, still lags behind the near-instantaneous, zero-shot reasoning of human solvers. These findings suggest that while \model is a promising step toward visual intelligence, current architectures still lack the innate spatial inductive biases of humans.\looseness=-1

In summary, our contributions are the following:

\begin{itemize}[leftmargin=*]
    \item We present \model, an editing-as-reasoning framework for visual reasoning.

    \item We introduce \bench, an abstract visual planning benchmark that covers two complementary planning forms, alongside automatic metrics for both pixel-wise fidelity and logical validity.

    \item We design controlled experiments to systematically probe intrinsic visual planning across a diverse suite of image editing models.

    \item We provide an in-depth analysis of the generalizability, scaling effect, and efficiency gap between neural visual planning and human solvers.
\end{itemize}

%

\section{The \bench Benchmark}

We propose the \bench benchmark, which consists of the classical Maze and Queen puzzles for assessing and analyzing intrinsic visual planning of image-editing models. The reasons that we choose the Maze and Queen tasks as the testbed are three-fold. First, they respectively cover two complementary paradigms of visual planning: locally-constrained sequential planning and globally-constrained combinatorial planning.
Second, they minimize visual recognition complexity—comprising primarily abstract structures—thus allowing for isolating the visual planning ability from multimodal understanding dependencies (\S\ref{sec:data-curation}). Third, unlike VLM-based evaluation that focuses more on qualitative assessment, they admit automatic metrics to accurately quantify logical correctness (\S\ref{sec:auto-metric}). 

\subsection{Automatic Data Curation} \label{sec:data-curation}

We generate both the Maze and Queen tasks procedurally. The complexity of a task is primarily defined by its scale, which ranges from $3\times3$ to $16\times16$ for Maze and $4\times4$ to $10\times10$ for Queen. The lower and upper bounds on the scale are chosen to avoid trivial solutions while allowing for efficient task generation. In Maze task, we additionally vary the geometry type to cover circular, hexagonal, square, and triangular geometries~\citep{codebox_mazes}, enabling fine-grained analysis. For each combination of maze scale and solution algorithm (including both depth-first and breadth-first search), we generate 50 mazes, totaling 2,800 test examples, that is, 700 per geometry type. For Queen, we randomly sample 50 puzzles per scale, resulting in a total of 350 test examples.

\subsection{Automatic Evaluation Metrics} \label{sec:auto-metric}
A fundamental challenge in generative image tasks is that high-quality visual outputs do not necessarily correspond to the right plan. Traditional metrics such as VLM-based critics~\citep{unifiedreward} and fidelity-oriented metrics~\citep{FID, LPIPS}) are inadequate for assessing the logical correctness of visual planning. Since \bench is generated procedurally, it enables rule-based metrics that automatically evaluate the correctness of generated plans. Concretely, we define logical validity as the following:

\begin{definitionbox}
\textbf{Logical validity} measures whether the generated solution matches the goal solution at the cell level. We compute a \textsc{Coverage} ratio, which measures the proportion of the goal solution that is correctly generated, and a \textsc{Violation} ratio, which measures the proportion of the generated solution that deviates from the goal solution. We further define \textsc{Pass} as $\max(0, \text{\textsc{Coverage}} - \text{\textsc{Violation}})$. \textsc{Pass} = 1 means the generated solution matches the solution structure exactly.
\end{definitionbox}

We further complement  logical validity with pixel-wise fidelity, defined as:
\begin{definitionbox}
 \textbf{Pixel-wise fidelity} measures pixel-wise differences that are measured using the mean squared error (MSE) between the generated and ground-truth images. We compute it separately for the solution area (cells covered by the goal solution, indicated by \textsc{Mse-In}) and the non-solution area  (indicated by \textsc{Mse-Out}).
\end{definitionbox}

\subsection{Consensus with Human Judges} 

To validate the reliability of our proposed automatic metric for logical validity, we measure the agreement between it and human judges. Specifically, we randomly sample 50 images per task for each evaluated model. Three human annotators were tasked with binary classification: check whether the generated solution successfully matches the ground truth image without any violations to the naked eye. We compare the results of human judges against our \textsc{Pass} rate; the agreement rate is 98\%, implying the high reliability of our automatic metric. Regarding the 2\% discrepancy, we find that it primarily arises from two scenarios: (1) complex tasks that cause human perception errors; and (2) overly faint solutions or altered non-solution areas. Fortunately, our automatic metric for pixel-wise fidelity helps detect these failure cases.

%

\begin{table*}[t!]\small
\vspace{-20pt} 
\renewcommand{\arraystretch}{1.2}
\setlength{\tabcolsep}{3pt}

\ifcsname ver@neurips_2026.sty\endcsname
\captionsetup{position=top,skip=7pt}
\caption{
Main results (\%) on \bench, including the Maze and Queen tasks. $\downarrow$ indicates lower is better, while $\uparrow$ indicates higher is better.
}
\label{tab:main}
\fi
\resizebox{\textwidth}{!}{%
\begin{tabular}{l@{\hspace{3pt}} cccccc @{\hspace{20pt}} cccccc}
\toprule
\multirow{2}{*}{\textbf{Model}} 
& \multicolumn{6}{c}{\textbf{Continuous (Maze) Task}} 
& \multicolumn{6}{c}{\textbf{Discrete (Queen) Task}} \\

\cmidrule(lr{20pt}){2-7} \cmidrule(lr){8-13}

& \textbf{Violation}$\downarrow$
& \textbf{Coverage}$\uparrow$
& \textbf{MSE In}$\downarrow$
& \textbf{MSE Out}$\downarrow$
& \textbf{Pass@1}$\uparrow$
& \textbf{Pass@5}$\uparrow$

& \textbf{Violation}$\downarrow$
& \textbf{Coverage}$\uparrow$
& \textbf{MSE In}$\downarrow$
& \textbf{MSE Out}$\downarrow$
& \textbf{Pass@1}$\uparrow$
& \textbf{Pass@5}$\uparrow$ \\

\midrule
\rowcolor{groupgray}
\multicolumn{13}{c}{\textit{proprietary models}} \\
\midrule 

GPT-image-1  
& \cellcolor{yellowLow} 62.88 
& \cellcolor{yellowMid} 58.97 
& \cellcolor{greenLow} 41.16 
& \cellcolor{greenLow} 52.76 
& \cellcolor{blueLow} 5.40 
& \cellcolor{blueLow} 6.06 
& \cellcolor{yellowLow} 62.91 
& \cellcolor{yellowLow} 37.09 
& \cellcolor{greenLow} 11.84 
& \cellcolor{greenLow} 5.87 
& \cellcolor{blueLow} 0.00 
& \cellcolor{blueLow} 2.28 \\

NanoBanana-Pro
& \cellcolor{yellowLow} 47.76 
& \cellcolor{yellowHigh} \textbf{64.21} 
& \cellcolor{greenLow} 24.20 
& \cellcolor{greenLow} 17.21 
& \cellcolor{blueLow} 4.82 
& \cellcolor{blueLow} 9.28 
& \cellcolor{yellowLow} 32.56 
& \cellcolor{yellowLow} 67.43 
& \cellcolor{greenLow} 9.10 
& \cellcolor{greenLow} 1.62 
& \cellcolor{blueHigh} \textbf{30.35} 
& \cellcolor{blueHigh} \textbf{35.58} \\

Seedream-4.5  
& \cellcolor{yellowLow} 16.90 
& \cellcolor{yellowLow} 25.67 
& \cellcolor{greenLow} 28.82 
& \cellcolor{greenLow} 30.96 
& \cellcolor{blueLow} 2.14 
& \cellcolor{blueLow} 3.21 
& \cellcolor{yellowLow} 76.86 
& \cellcolor{yellowLow} 23.14 
& \cellcolor{greenLow} 11.55 
& \cellcolor{greenLow} 5.95 
& \cellcolor{blueLow} 2.86 
& \cellcolor{blueLow} 2.86 \\

\midrule
\rowcolor{groupgray}
\multicolumn{13}{c}{\textit{open-source models (w/o chain-of-thought reasoning)}} \\
\midrule

Flux-Kontext-Dev 
& \cellcolor{yellowLow} 23.84 
& \cellcolor{yellowLow} 30.24 
& \cellcolor{greenLow} 30.96 
& \cellcolor{greenLow} 18.31 
& \cellcolor{blueLow} 0.36 
& \cellcolor{blueLow} 3.57 
& \cellcolor{yellowLow} 78.63 
& \cellcolor{yellowLow} 21.37 
& \cellcolor{greenLow} 11.48 
& \cellcolor{greenLow} 7.71 
& \cellcolor{blueLow} 0.92 
& \cellcolor{blueLow} 2.34 \\

Qwen-Image-Edit 
& \cellcolor{yellowLow} 19.37 
& \cellcolor{yellowLow} 28.51 
& \cellcolor{greenLow} 18.82 
& \cellcolor{greenLow} 5.70 
& \cellcolor{blueLow} 1.43 
& \cellcolor{blueLow} 2.14 
& \cellcolor{yellowLow} 69.52 
&\cellcolor{yellowLow}  30.47 
& \cellcolor{greenLow} 8.83 
& \cellcolor{greenLow} 5.30 
& \cellcolor{blueLow} 2.86 
& \cellcolor{blueLow} 4.00 \\

Bagel 
& \cellcolor{yellowLow} 28.91 
& \cellcolor{yellowLow} 27.15 
& \cellcolor{greenLow} 11.64 
& \cellcolor{greenLow} 5.84 
& \cellcolor{blueLow} 0.00 
& \cellcolor{blueLow} 1.00 
& \cellcolor{yellowLow} 61.57 
& \cellcolor{yellowLow} 38.43 
& \cellcolor{greenLow} 8.94 
& \cellcolor{greenLow} 1.22 
& \cellcolor{blueLow} 0.00 
& \cellcolor{blueLow} 0.00 \\

Janus-Pro 
& \cellcolor{yellowHigh} \textbf{5.41} 
& \cellcolor{yellowLow} 1.85 
& \cellcolor{greenLow} 57.47 
& \cellcolor{greenLow} 76.80 
& \cellcolor{blueLow} 0.00 
& \cellcolor{blueLow} 0.00 
& \cellcolor{yellowLow} 84.24 
& \cellcolor{yellowLow} 15.76 
& \cellcolor{greenLow} 12.97 
& \cellcolor{greenLow} 9.83 
& \cellcolor{blueLow} 0.00 
& \cellcolor{blueLow} 0.57 \\

\midrule

Bagel (fine-tuned)  
& \cellcolor{yellowLow} 12.21 
& \cellcolor{yellowLow} 51.02 
& \cellcolor{greenHigh} \textbf{8.66} 
& \cellcolor{greenHigh} \textbf{3.07} 
& \cellcolor{blueMid} 11.54 
& \cellcolor{blueHigh} \textbf{23.64}
& \cellcolor{yellowLow} 68.27 
& \cellcolor{yellowLow} 31.73 
& \cellcolor{greenHigh} \textbf{6.05} 
& \cellcolor{greenHigh} \textbf{0.63} 
& \cellcolor{blueMid} 14.29
& \cellcolor{blueMid} 14.57 \\

Janus-Pro (fine-tuned) 
& \cellcolor{yellowLow} 35.60 
& \cellcolor{yellowLow} 23.33 
& \cellcolor{greenLow} 55.99 
& \cellcolor{greenLow} 50.94 
& \cellcolor{blueLow} 1.43 
& \cellcolor{blueLow} 2.22  
& \cellcolor{yellowHigh} \textbf{16.07} 
& \cellcolor{yellowHigh} \textbf{83.93} 
& \cellcolor{greenLow} 7.91 
& \cellcolor{greenLow} 1.38 
& \cellcolor{blueLow} 12.57 
& \cellcolor{blueLow} 13.03 \\

\midrule
\rowcolor{groupgray}
\multicolumn{13}{c}{\textit{w/ chain-of-thought reasoning}} \\
\midrule

Bagel
& \cellcolor{yellowLow} 34.06 
& \cellcolor{yellowLow} 30.31
& \cellcolor{greenLow} 14.77 
& \cellcolor{greenMid} 3.97 
& \cellcolor{blueLow} 0.00 
& \cellcolor{blueLow} 0.57 
& \cellcolor{yellowLow} 98.41 
& \cellcolor{yellowLow} 1.59 
& \cellcolor{greenLow} 9.63 
& \cellcolor{greenLow} 1.40 
& \cellcolor{blueLow} 0.00 
& \cellcolor{blueLow} 0.00 \\

Bagel (fine-tuned) 
& \cellcolor{yellowLow} 15.24 
& \cellcolor{yellowLow} 44.65 
& \cellcolor{greenMid} 10.17 
& \cellcolor{greenLow} 5.25 
& \cellcolor{blueHigh} \textbf{17.90} 
& \cellcolor{blueMid} 18.42  
& \cellcolor{yellowLow} 64.22 
& \cellcolor{yellowLow} 35.78 
& \cellcolor{greenMid} 6.13 
& \cellcolor{greenMid} 0.72 
& \cellcolor{blueLow} 14.08 
& \cellcolor{blueLow} 14.11 \\

Janus-Pro
& \cellcolor{yellowMid} 6.03 
& \cellcolor{yellowLow} 0.89 
& \cellcolor{greenLow} 53.02 
& \cellcolor{greenLow} 73.98 
& \cellcolor{blueLow} 0.00 
& \cellcolor{blueLow} 0.00 
& \cellcolor{yellowLow} 82.91 
& \cellcolor{yellowLow} 17.09 
& \cellcolor{greenLow} 10.93 
& \cellcolor{greenLow} 8.04 
& \cellcolor{blueLow} 0.00 
& \cellcolor{blueLow} 0.70 \\

Janus-Pro (fine-tuned) 
& \cellcolor{yellowLow} 31.23 
& \cellcolor{yellowLow} 25.12 
& \cellcolor{greenLow} 56.81 
& \cellcolor{greenLow} 52.28 
& \cellcolor{blueLow} 2.79 
& \cellcolor{blueLow} 4.13  
& \cellcolor{yellowMid} 18.52 
& \cellcolor{yellowMid} 81.48 
& \cellcolor{greenLow} 6.48 
& \cellcolor{greenLow} 1.67 
& \cellcolor{blueLow} 11.20 
& \cellcolor{blueLow} 13.56 \\

\bottomrule
\end{tabular}
}

\ifcsname ver@neurips_2026.sty\endcsname
\else
\caption{
Main results (\%) on \bench, including the Maze and Queen tasks. $\downarrow$ indicates lower is better, while $\uparrow$ indicates higher is better.
}
\label{tab:main}
\fi
\end{table*}
\section{Experiment}

\subsection{Experimental Setup}

\paragraph{Evaluated models.} To investigate the intrinsic visual planning capabilities of current image editing models, we benchmark representative models from two dominant generative paradigms: diffusion-based and autoregressive models. To reveal the gaps between the proprietary and open-source domains, we consider the following image editing models:

\begin{itemize}[leftmargin=*]
    \item \textbf{Proprietary domain} includes frontier models like \text{GPT-Image-1}~\citep{gptimage1}, \text{NanoBanana-Pro}~\citep{gemini3pro_image_2025}\footnote{For NanoBanana-Pro, the use of Chain-of-Thought (CoT) is not publicly reported.}  and \text{Seedream-4.5}~\citep{seedream2025seedream40nextgenerationmultimodal}. 
    \item \textbf{Open-source domain} includes \text{Qwen-Image-Edit}~\citep{wu2025qwenimagetechnicalreport}, \text{Flux-Kontext-Dev}~\citep{labs2025flux1kontextflowmatching}, Bagel~\citep{deng2025bagel} and \text{Janus-Pro-7B}~\citep{chen2025janus}. Among them, \text{Qwen-Image-Edit}, \text{Flux-Kontext-Dev}, and \text{Bagel} are diffusion-based and \text{Janus-Pro-7B} is an autoregressive model.
\end{itemize}

\paragraph{Evaluation method.} We directly prompt models to draw out the required solution. We keep the prompt concise and clear and apply the same prompt to all models, minimizing variances arising from prompts (see the example evaluations in the Figure~\ref{fig:metric_example}). The complete prompts are provided in Appendix~\ref{app:instruction}.

\paragraph{Measures.} 
We evaluate each model 5 times and report the average \textsc{Pass@5}, \textsc{Mse-In}, \textsc{Mse-Out}, and \textsc{Coverage} and \textsc{Violation} ratios (see \S \ref{sec:auto-metric}). We also supplement with \textsc{Pass@1} using the first-round image generation.

\subsection{Main Results}

Our initial results show that Bagel and Janus-Pro struggle in the zero-shot setting, i.e., they fail to follow the instruction and generate valid solutions, likely because these are out-of-domain scenarios for them (see Table~\ref{tab:main}). Thus, to investigate their potential in acquiring the visual planning ability, we apply supervised fine-tuning. We curate a training set consisting of the simplest scale: $3\times3$ mazes spanning all four geometry types (circle, hexagon, square, and triangle) and 4-Queens puzzles. The training set comprises 800 samples per geometry type and 800 4-Queens puzzles, accompanied by a separate held-out set for validation. We train each model for up to 8 epochs, and apply early stopping when the MSE loss on the validation set plateaus. 

\paragraph{Frontier proprietary editing models have limited capacity in abstract visual planning.}
On Maze, proprietary editing models achieve the best \textsc{Pass@1} of 5.4\%, exhibiting limited zero-shot proficiency. They often fail to respect maze boundaries, generating paths that cut through walls. Among them, GPT-Image-1 exhibits the worst instruction-following capability, with a violation rate of 62.88\%. While NanoBanana-Pro performs best in terms of pixel-wise fidelity, it tends to over-generate paths that traverse the entire maze, indicated by its high violation rate (e.g., 47.76\%). Seedream-4.5 appears to respect the constraints ($<20\%$ violation), but this is through the shortcut of under-generation, i.e., it can hardly generate a complete path.
On Queen, while the best performing NanoBanana-Pro shows a high \textsc{Pass@1} of 30.35\%, all other proprietary models demonstrate nearly zero \textsc{Pass@1} in the zero-shot setting. The surprisingly high performance of NanoBanana-Pro indicates that it may have seen similar tasks during training.

\begin{figure*}[t!]\small
\vspace{-20pt}
\centering
    \begin{subfigure}[t]{0.15\textwidth}
        \centering
        \includegraphics[width=\linewidth]{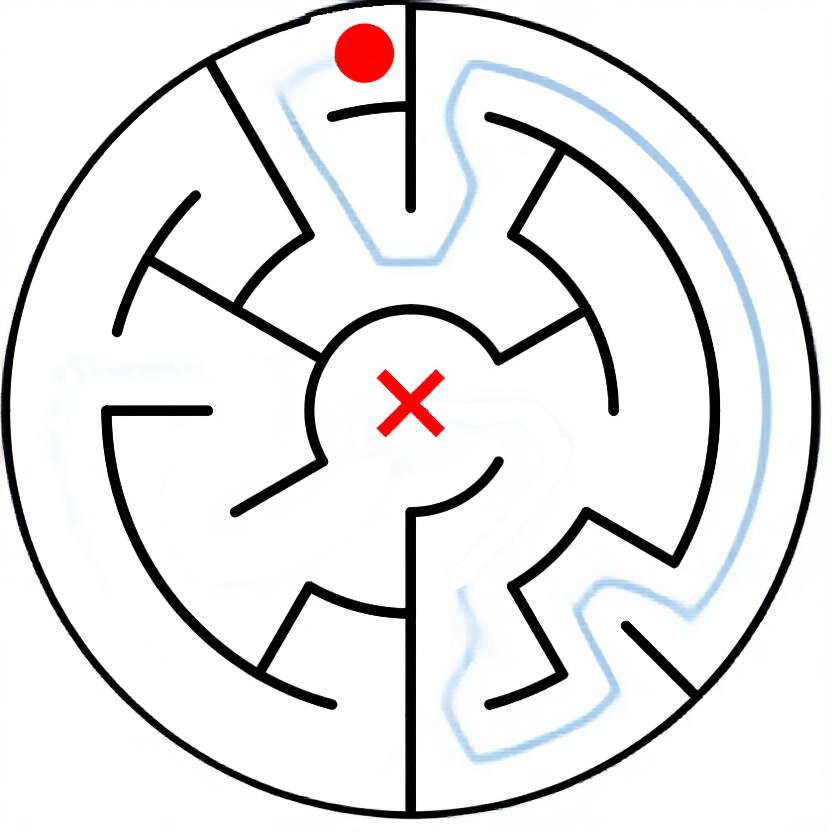}
    \end{subfigure}
    \hfill 
    \begin{subfigure}[t]{0.15\textwidth}
        \centering
        \includegraphics[width=\linewidth]{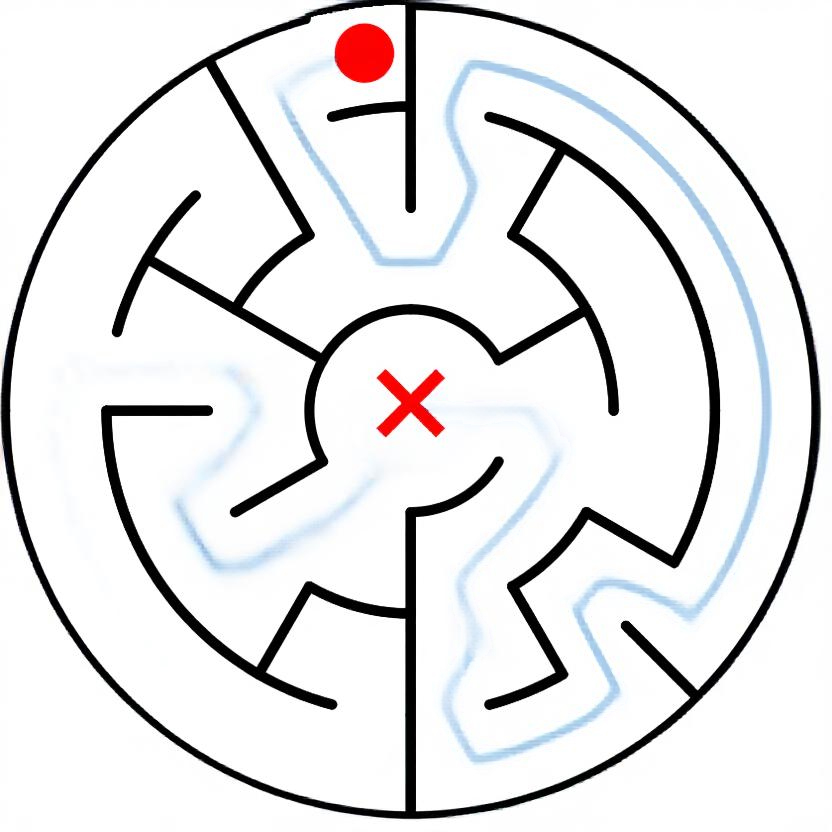}
    \end{subfigure}
    \hfill
    \begin{subfigure}[t]{0.15\textwidth}
        \centering
        \includegraphics[width=\linewidth]{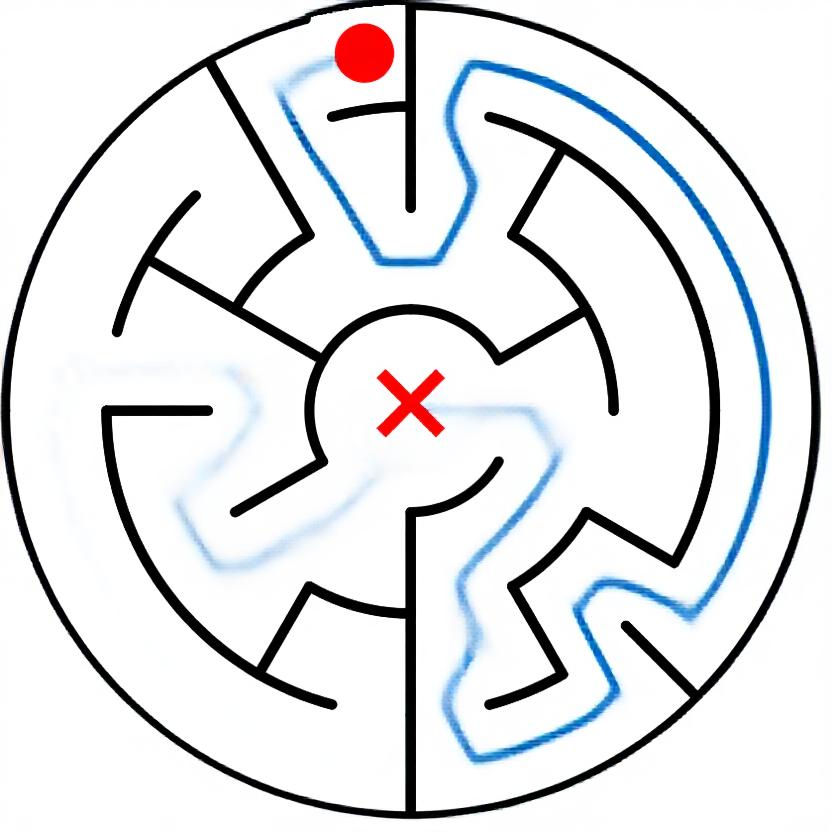}
    \end{subfigure}
    \hfill
    \begin{subfigure}[t]{0.15\textwidth}
        \centering
        \includegraphics[width=\linewidth]{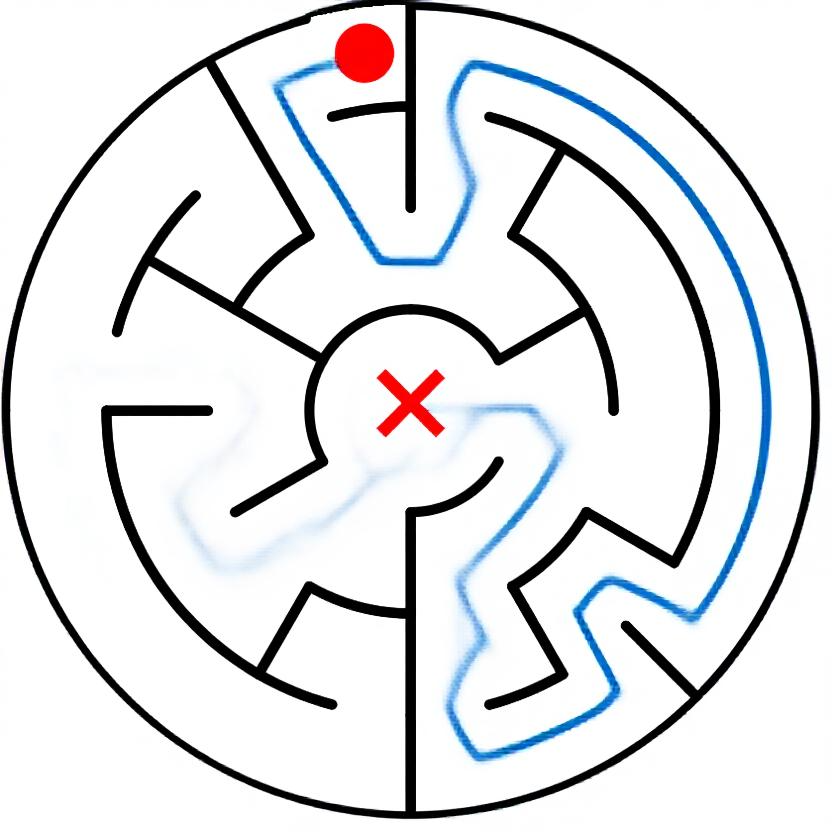}
    \end{subfigure}
    \hfill
    \begin{subfigure}[t]{0.15\textwidth}
        \centering
        \includegraphics[width=\linewidth]{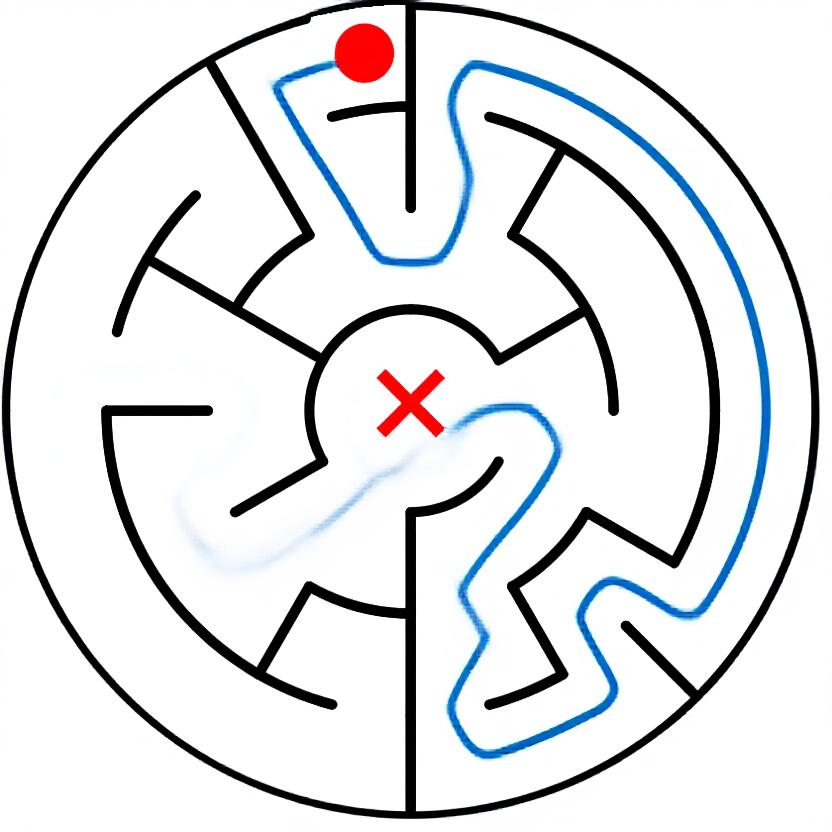}
    \end{subfigure}
    \hfill
    \begin{subfigure}[t]{0.15\textwidth}
        \centering
        \includegraphics[width=\linewidth]{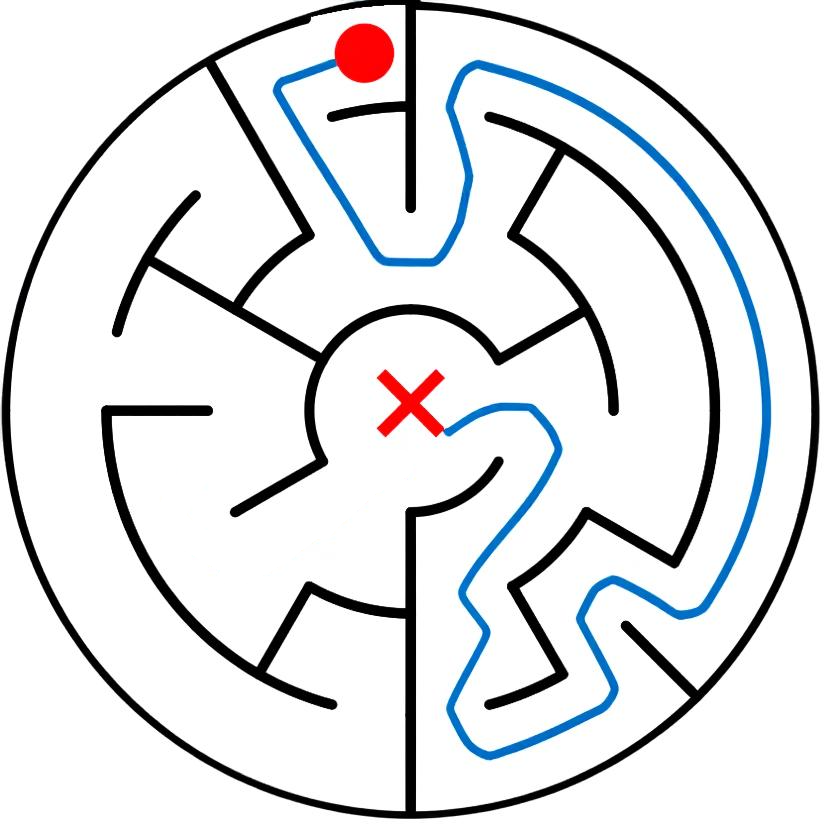}
    \end{subfigure}

    \par\smallskip 

    \begin{subfigure}[t]{0.15\textwidth}
        \centering
        \includegraphics[width=\linewidth]{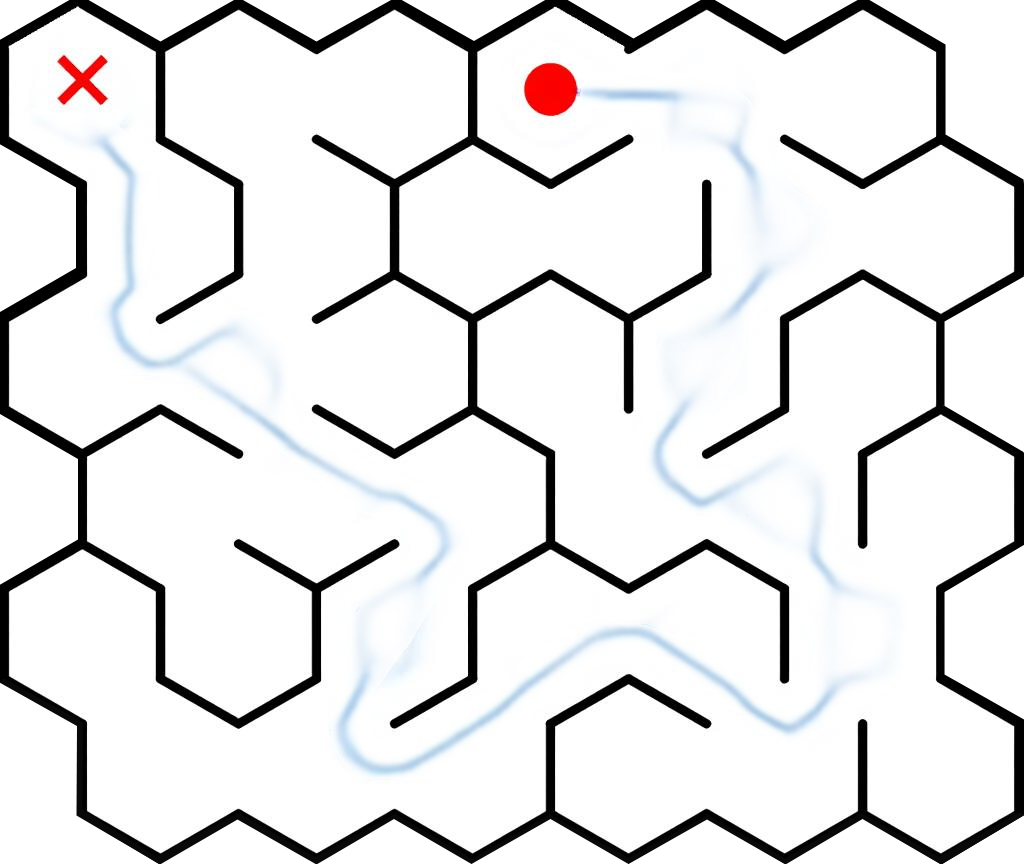}
    \end{subfigure}
    \hfill
    \begin{subfigure}[t]{0.15\textwidth}
        \centering
        \includegraphics[width=\linewidth]{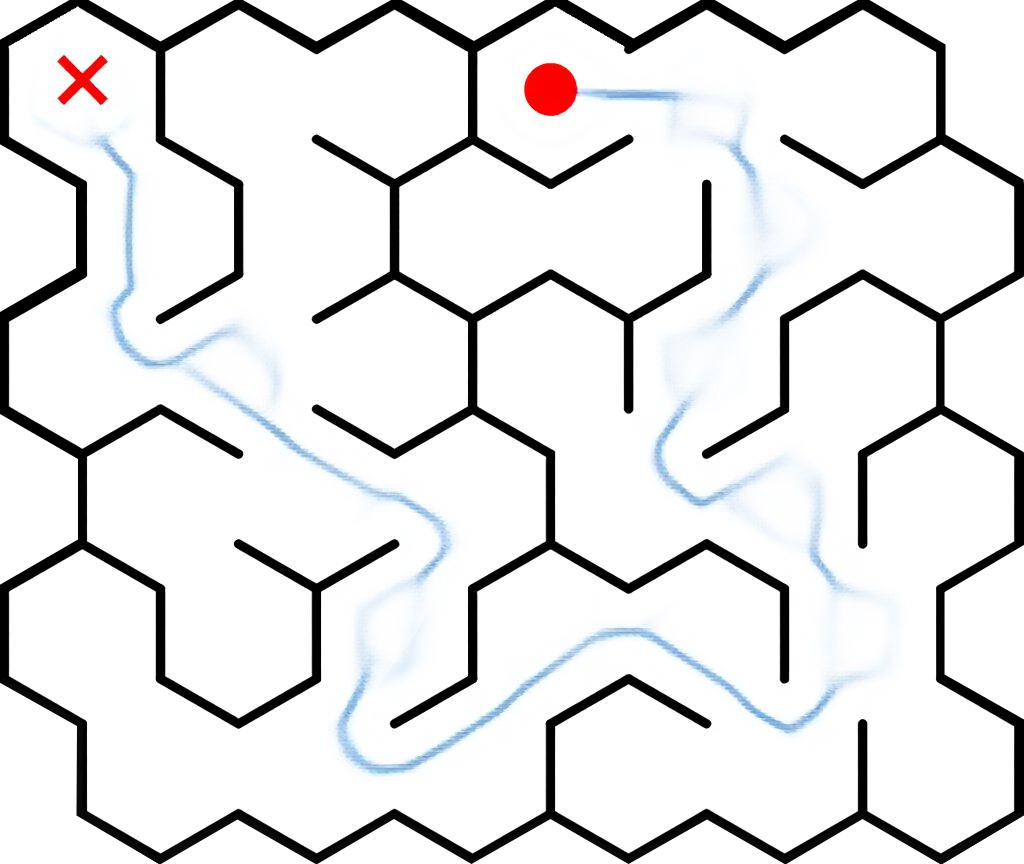}
    \end{subfigure}
    \hfill
    \begin{subfigure}[t]{0.15\textwidth}
        \centering
        \includegraphics[width=\linewidth]{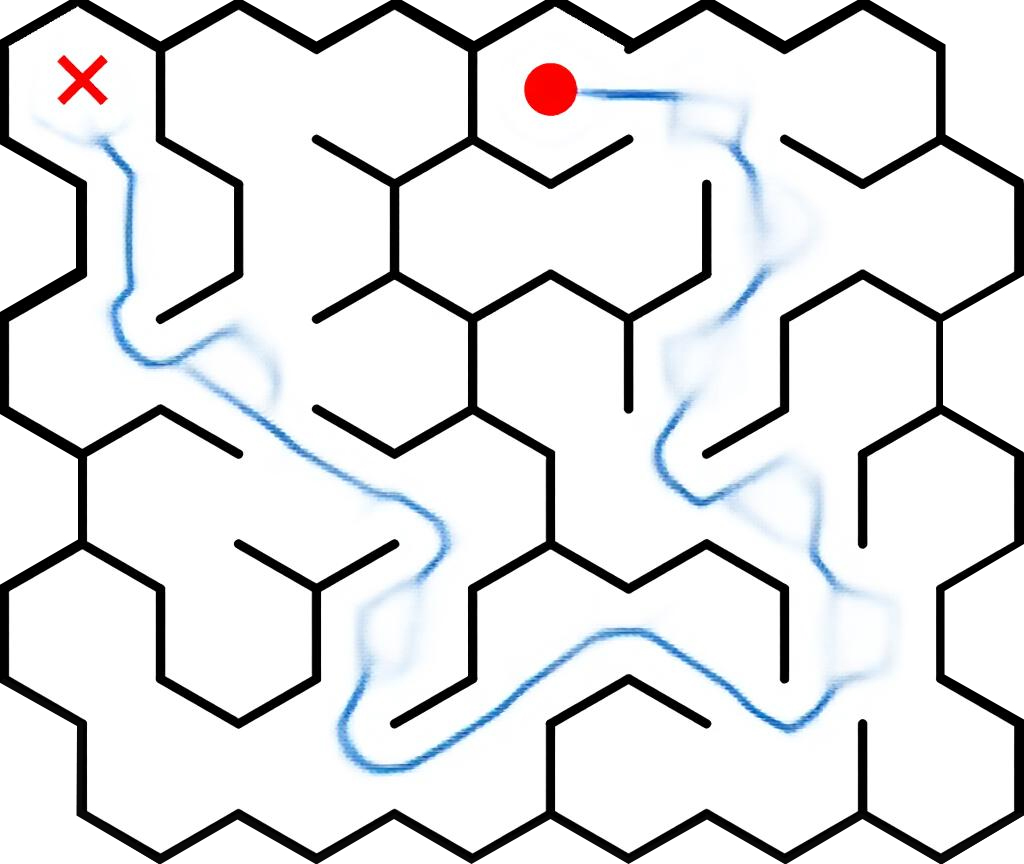}
    \end{subfigure}
    \hfill
    \begin{subfigure}[t]{0.15\textwidth}
        \centering
        \includegraphics[width=\linewidth]{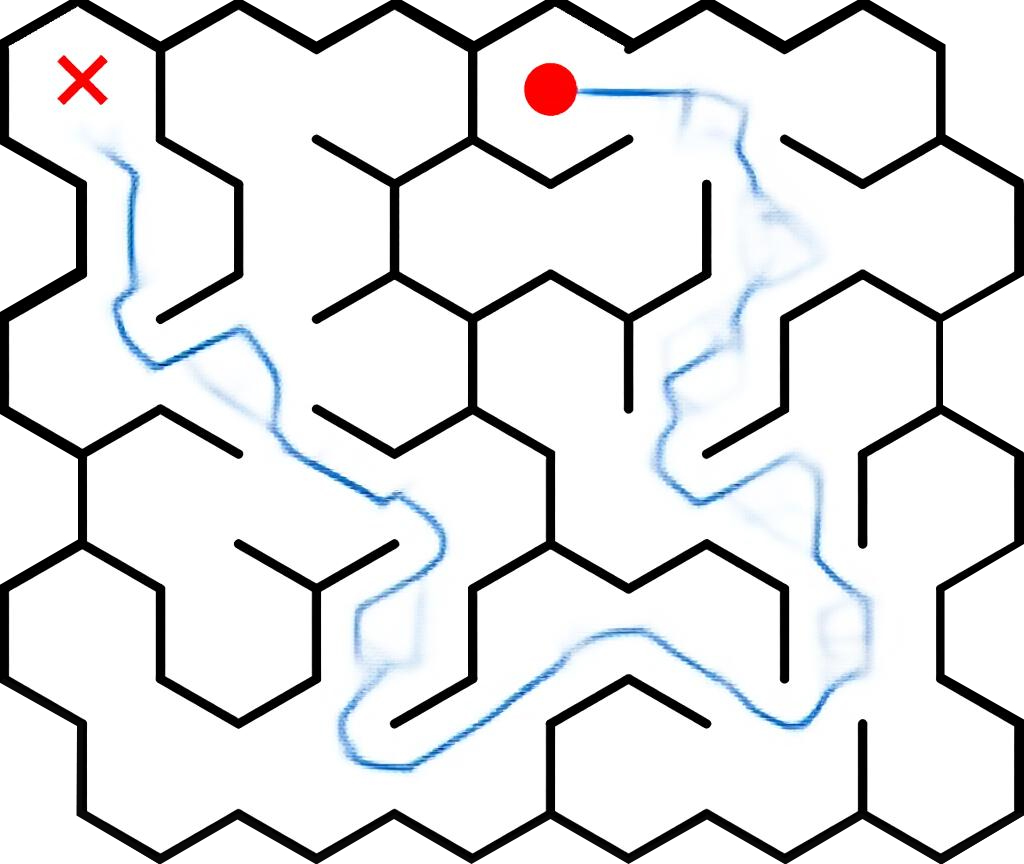}
    \end{subfigure}
    \hfill
    \begin{subfigure}[t]{0.15\textwidth}
        \centering
        \includegraphics[width=\linewidth]{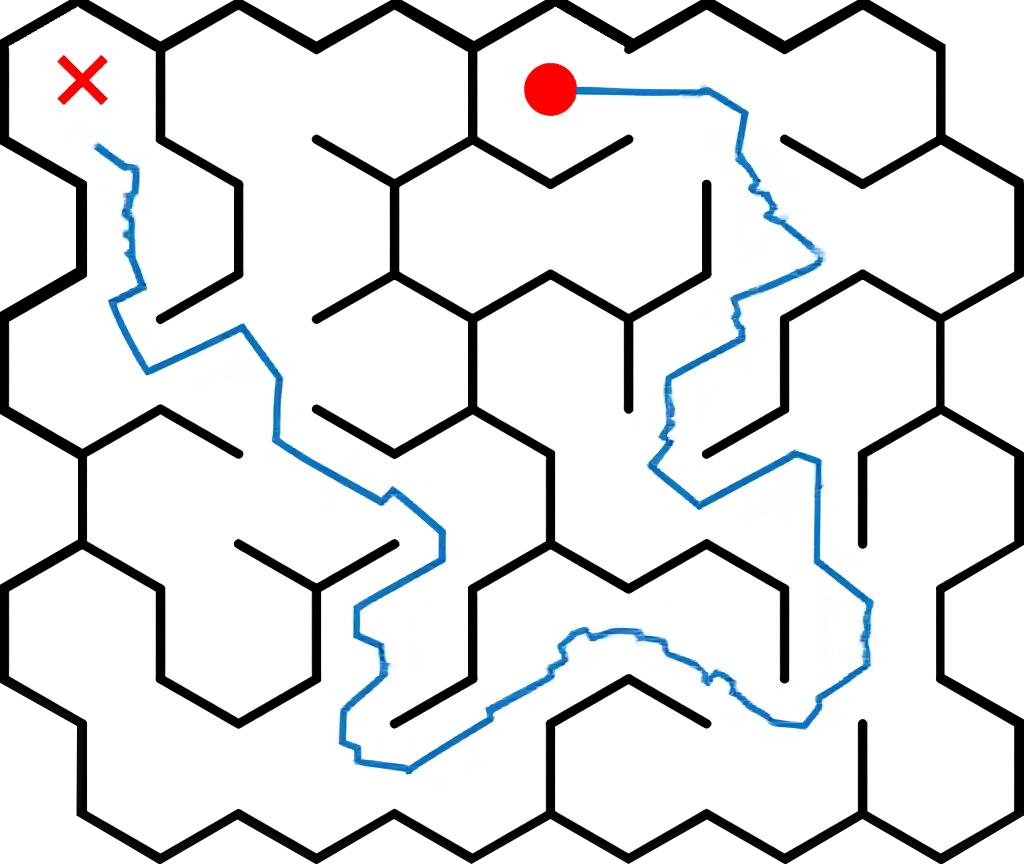}
    \end{subfigure}
    \hfill
    \begin{subfigure}[t]{0.15\textwidth}
        \centering
        \includegraphics[width=\linewidth]{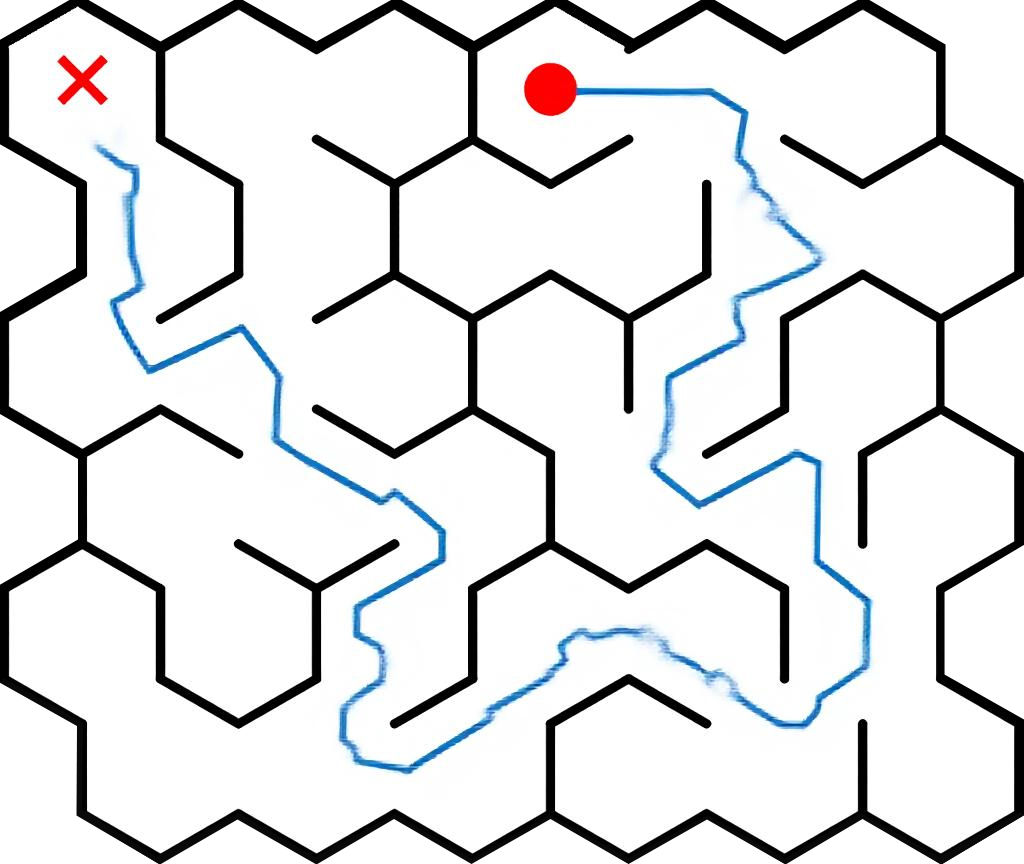}
    \end{subfigure}

    \par\smallskip 

     \begin{subfigure}[t]{0.15\textwidth}
        \centering
        \includegraphics[width=\linewidth]{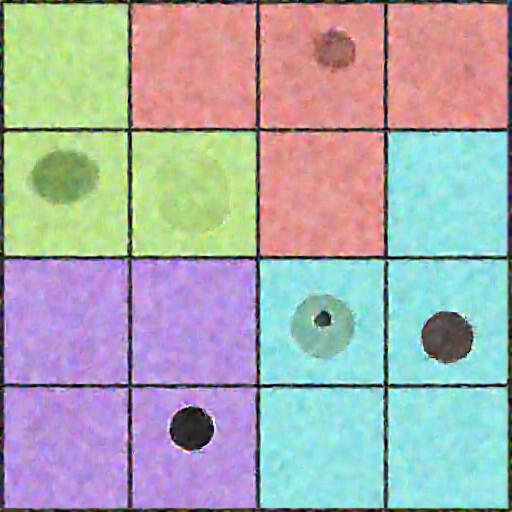}
        \subcaption{$t=1$}
    \end{subfigure}
    \hfill
    \begin{subfigure}[t]{0.15\textwidth}
        \centering
        \includegraphics[width=\linewidth]{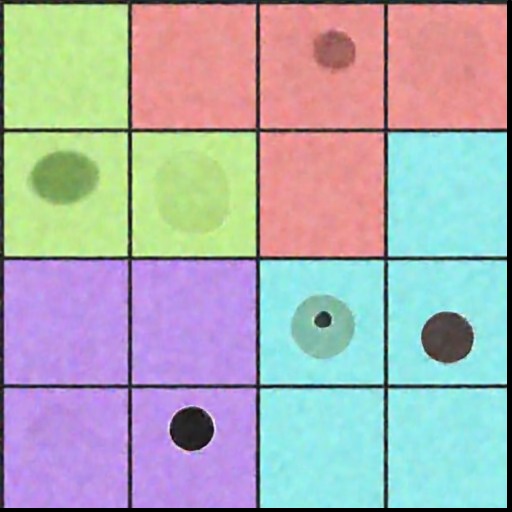}
        \subcaption{$t=2$}
    \end{subfigure}
    \hfill
    \begin{subfigure}[t]{0.15\textwidth}
        \centering
        \includegraphics[width=\linewidth]{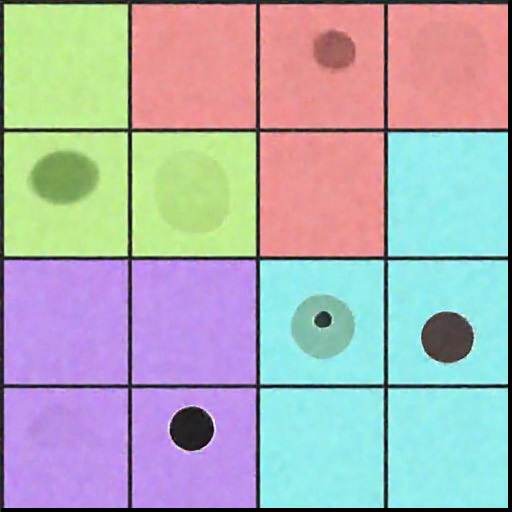}
        \subcaption{$t=4$}
    \end{subfigure}
    \hfill
    \begin{subfigure}[t]{0.15\textwidth}
        \centering
        \includegraphics[width=\linewidth]{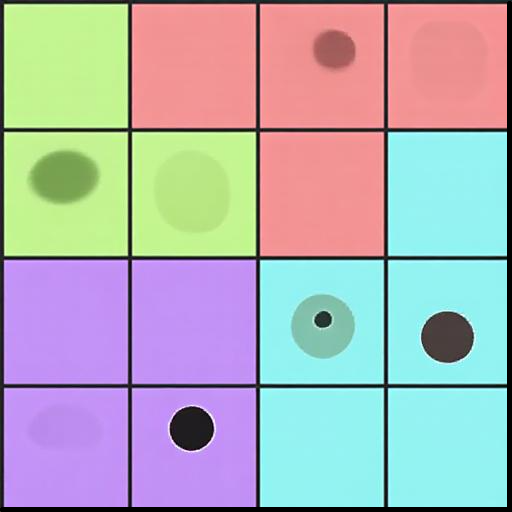}
        \subcaption{$t=6$}
    \end{subfigure}
    \hfill
    \begin{subfigure}[t]{0.15\textwidth}
        \centering
        \includegraphics[width=\linewidth]{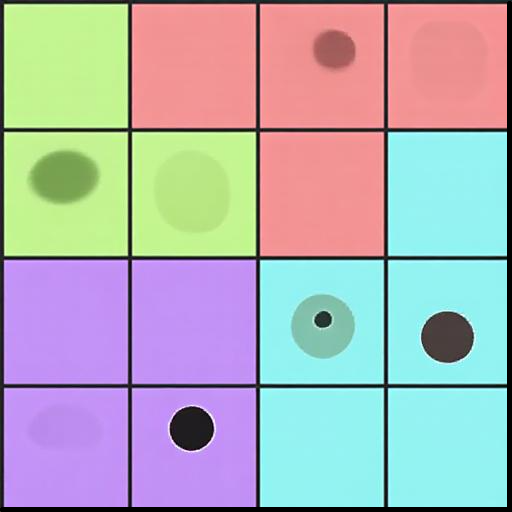}
        \subcaption{$t=8$}
    \end{subfigure}
    \hfill
    \begin{subfigure}[t]{0.15\textwidth}
        \centering
        \includegraphics[width=\linewidth]{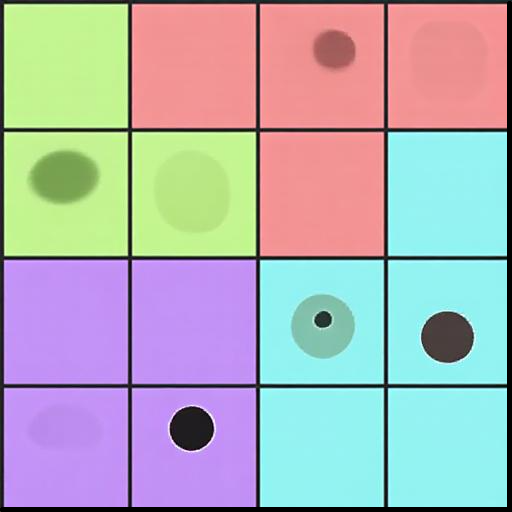}
        \subcaption{$t=10$}
    \end{subfigure}

    \caption{Solutions from different denoising steps ($t$) of a fine-tuned Bagel on Maze ($\iscircle$ and $\ishexagon$) and Queen (last row) task. 
    }
    \label{fig:diffusion_cot}
    \vspace{-12pt} 
\end{figure*}

\paragraph{Diffusion-based models may be more effective at developing visual reasoning logic than autoregressive models.}
We analyze which learning paradigm is better at developing visual reasoning logic. To do so, we compare Bagel~\citep{deng2025bagel} and Janus-Pro~\citep{chen2025janus}, two representatives of image editing models, respectively. Without fine-tuning, both have a zero \textsc{Pass@1}; after fine-tuning, Bagel improves \textsc{Pass@1} from 0 to 11.54\% on Maze, but the performance of Janus-Pro is only 1.43\%. A similar trend is observed on Queen: the fine-tuned Bagel achieves 14.29\% \textsc{Pass@1}, while the fine-tuned Janus-Pro lags, reaching only 12.57\%.
Though the lack of transparency regarding training precludes a definitive conclusion, these findings suggest that diffusion-based modeling may be more effective at developing visual reasoning logic. We hypothesize that the progressive denoising in diffusion models fosters a global structural awareness that is beneficial for visual planning. Conversely, the sequential, token-based nature of autoregressive models lacks this global perspective, as generation is constrained by a local, raster-scan order.

\paragraph{Chain-of-Thought prompting is not always helpful.}
We further evaluate the models using Chain-of-Thought (CoT) prompting, but the results are mixed. For unified multimodal architectures such as Bagel and Janus-Pro, CoT provides negligible benefits in the zero-shot regime. However, it yields marginal improvements following fine-tuning, suggesting that the models must first internalize the task's underlying logic before they can effectively leverage intermediate reasoning steps.

\paragraph{Qualitative studies of ``visual planning".} 
We provide a qualitative study of the ``planning" process of a fine-tuned Bagel on both Maze and Queen tasks (see Figure~\ref{fig:diffusion_cot}, more examples are shown in Appendix~\ref{app:qualitative_more}). 
On Maze, the model exhibits a clear global-planning behavior. The overall solution path emerges at early denoising steps (e.g., $t=1,2,4$) with low confidence, indicated by faint trajectories, and is progressively refined over time. Incorrect subpaths are gradually corrected (e.g., $t=8$), leading to a valid solution at later steps (e.g., $t=10$). This coarse-to-fine trajectory construction aligns with the denoising nature of diffusion models, where the global structure is iteratively improved.
On Queen, we observe a distinct planning pattern: a coarse global configuration of placements is established in the initial steps, followed by fine-grained adjustments. This contrast highlights the differences between the two paradigms. While sequential tasks like Maze are amenable to iterative and local refinements, combinatorial tasks like Queen necessitate significant global updates. Such global coordination remains a formidable challenge for current editing models.

\subsection{Generalizability}

We further investigate how well editing models can generalize to unseen geometry types and scales. For this study, we use a fine-tuned Bagel as it demonstrates non-trivial visual planning capabilities. For Maze, we evaluate generalization across both geometry types and scales. The test set covers scales from $3\times3$ to $16\times16$, with 50 mazes sampled per scale for each geometry type. 
For Queen, we evaluate across scales from $4\times4$ to $10\times10$, with 50 samples per scale.

\subsubsection{Cross-Geometry Generalization}

\paragraph{Fine-tuning on $\ishexagon$ yields the best generalization across other geometry types.}

\begin{wrapfigure}{r}{0.6\linewidth}
    \centering
    \vspace{-15pt} 
    \includegraphics[width=\linewidth]{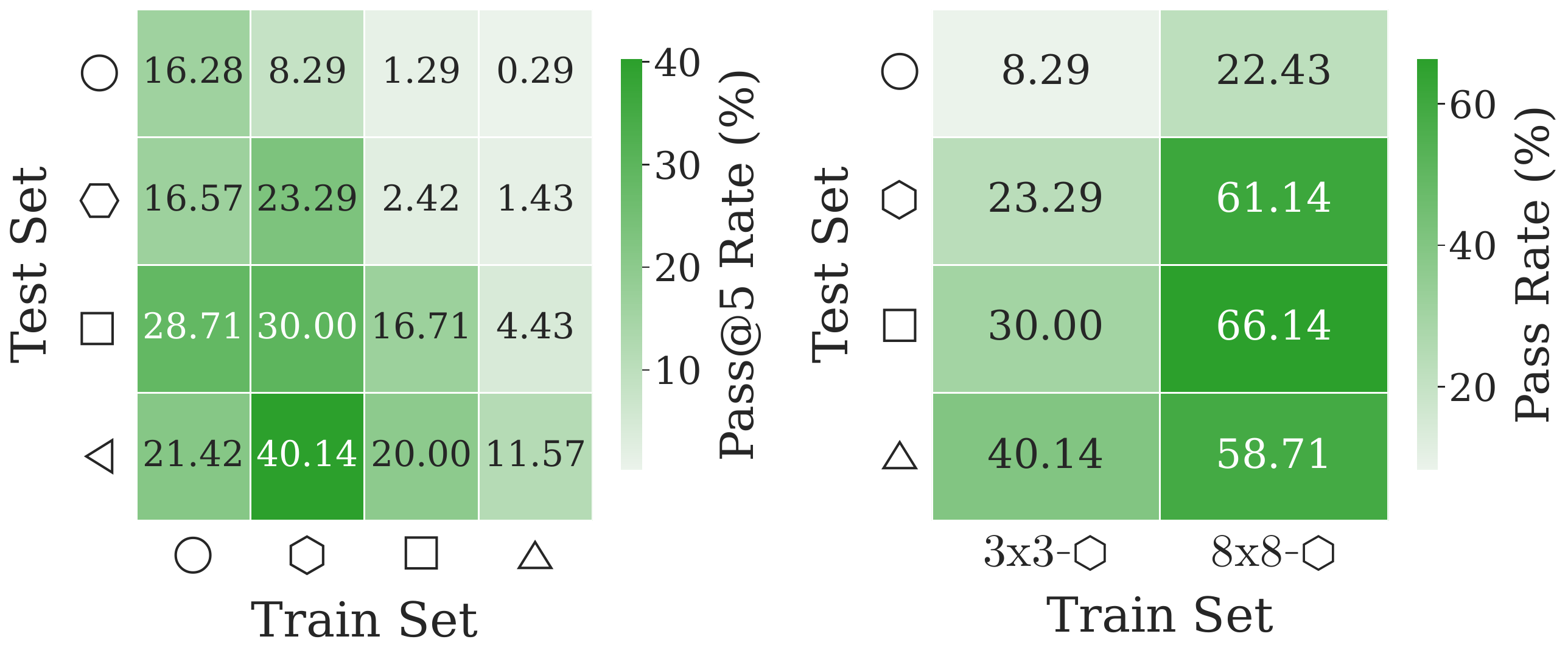} 
    \caption{\small Zero-shot generalization. \textbf{Left:} \textsc{Pass@5} matrix for $3\times3$ models. \textbf{Right:} Comparison between $3\times3$ and $8\times8$ $\ishexagon$ training.}
    \label{fig:generalization} 
    \vspace{-10pt} 
\end{wrapfigure}
We evaluate Bagel’s zero-shot generalization across geometry types (See Figure~\ref{fig:generalization} (left)), revealing an asymmetric transfer pattern: training on complex geometries (e.g., hexagons) yields better performance on simpler ones than vice-versa. Notably, the hexagon-trained model generalizes best—achieving 40.14\% on triangles and 30.00\% on squares, outperforming in-domain baselines. 
We attribute this to the more variable directions in hexagonal mazes. Their action space functions as a superset that encompasses the geometric constraints of both square and triangular mazes. This suggests that the models have learned fundamental path-finding logic that transcends specific geometries. 

\paragraph{Fine-tuning on larger-scale $\ishexagon$ enhances cross-geometry generalization. }
To further explore if increased training complexity reinforces cross-geometry generalization, we extend our study to a $8\times8$ training setting. As illustrated in Figure~\ref{fig:generalization} (right), increasing the scale of the training mazes leads to a substantial leap in generalization performance across all test domains, which indicates that exposure to larger-scale problems forces the model to transition from learning in-domain geometric patterns to out-of-domain visual planning capabilities.

\subsubsection{Cross-Scale Generalization}

\paragraph{Fine-tuning on $3\times 3$ mazes yields generalization to larger scales up to $16\times 16$.}
We further investigate cross-scale generalization. For this analysis, we fine-tuned Bagel on $\ishexagon$ mazes since they induce the best cross-geometry generalization. Surprisingly, fine-tuning on simple $3\times 3$ mazes enables generalization to larger scales up to $16\times 16$ (see Figure~\ref{fig:hex_queen}). We again extend our study to the $8\times 8$ training setting. Unsurprisingly, more complex training mazes lead to better zero-shot cross-scale generalization. However, while $8\times8$ trained model excels at maintaining local structural constraints, indicated by the low violation rate, it still struggles with the most complex mazes. We find that, when the scale increases, the model often generates perfect local paths near the starting and end points of the maze but fails to connect them in the middle, leading to near-zero success rate, presumably because the path length increases with the scale, making it more challenging for the model to maintain a growing long-distance dependency.

\begin{wrapfigure}{r}{0.7\linewidth}
    \centering
    \vspace{-20pt} 
    \includegraphics[width=\linewidth]{fig/hex_queen_figure.pdf} 
    \caption{Generalization across scales for Maze (top) and Queen (bottom) tasks. The dotted line and the solid line respectively represent baseline and trained model.}
    \label{fig:hex_queen}
    \vspace{-15pt} 
\end{wrapfigure}
\paragraph{Queen relies on more complex training scales for non-trivial cross-scale generalization.} As shown at the bottom of the Figure ~\ref{fig:hex_queen}, unlike Maze, which induces non-trivial cross-scale generalization when training from the smallest $3\times 3$ scale, fine-tuning on the smallest $4\times 4$ scale yields perfect in-domain performance but no generalization to larger scales, indicating a strong memorization. Consistent with our observations on Maze, fine-tuning on larger scales (e.g., $7\times7$) yields better, non-trivial cross-scale generalization. This suggests that for combinatorial visual planning, exposure to larger training scales is crucial to acquiring scale-invariant reasoning capabilities.

\subsection{Scaling Effect}\label{sec:scaling}
\begin{wrapfigure}{r}{0.4\linewidth}
    \centering
    \vspace{-20pt} 
    \includegraphics[width=\linewidth]{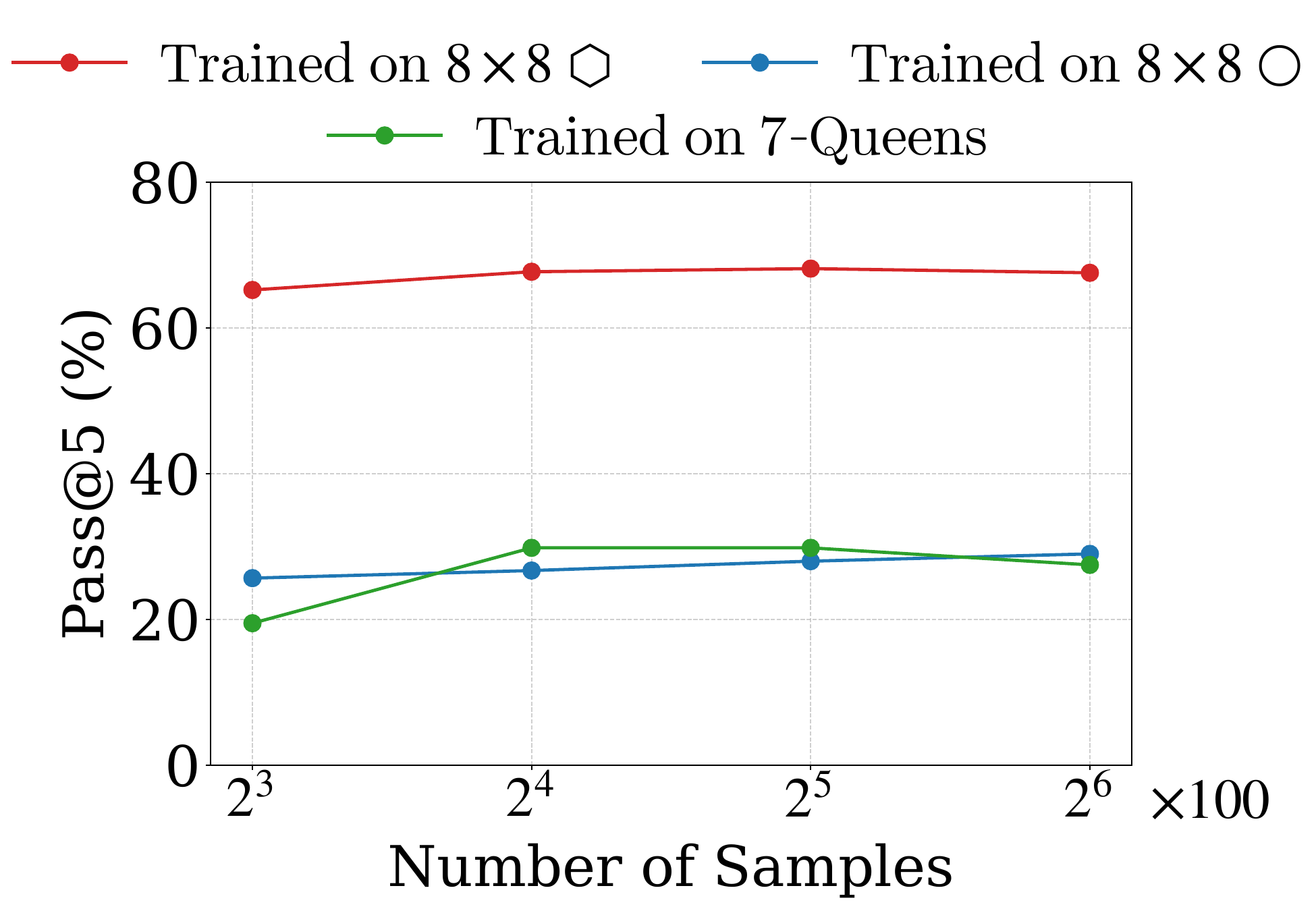} 
    \vspace{-15pt} 
    \caption{Data scaling.}
    \label{fig:data_scaling}
    \vspace{-10pt} 
\end{wrapfigure}

Next, we study if scaling up the training data and compute improves visual planning. For this analysis, we fine-tune Bagel on $8\times 8$ $\ishexagon$ mazes (representing the best performing geometry), $8\times 8$ $\iscircle$ mazes (representing the hardest geometry), and 7-Queens, respectively, and test on all scales of the same geometry type.

\paragraph{Scaling up training data.}

We analyze the effect of data scaling with $N \in \{800, 1600, 3200, 6400\}$ under a fixed compute budget of 1000 training steps. In general, scaling up training data initially yields slight improvements on all tasks, but the gains become marginal after $N>1600$ (see Figure~\ref{fig:data_scaling}).
On Maze, data scaling results in a quick performance saturation on both $\ishexagon$ and $\iscircle$ geometries, e.g., while the performance on $\ishexagon$ improves from 65.2\% to 68.1\% when increasing $N$ from 800 to 1600, it then stays plateaued, suggesting that scaling up training data mainly improves robustness to scale variation rather than the intrinsic sequential planning ability. 
Though the trend on 7-Queens is similar to that on Maze, scaling up training data from 800 to 1600 yields a much larger initial gain (+10.3\%), indicating that combinatorial tasks like Queen benefit a lot more from the highly diverse solution patterns.  
We also provide a more detailed analysis of data scaling on cross-domain geometries for Maze task in Appendix~\ref{app:data_scaling_ood}.  


\begin{wrapfigure}{r}{0.32\linewidth}
    \centering
    \vspace{-20pt} 
    \includegraphics[width=\linewidth]{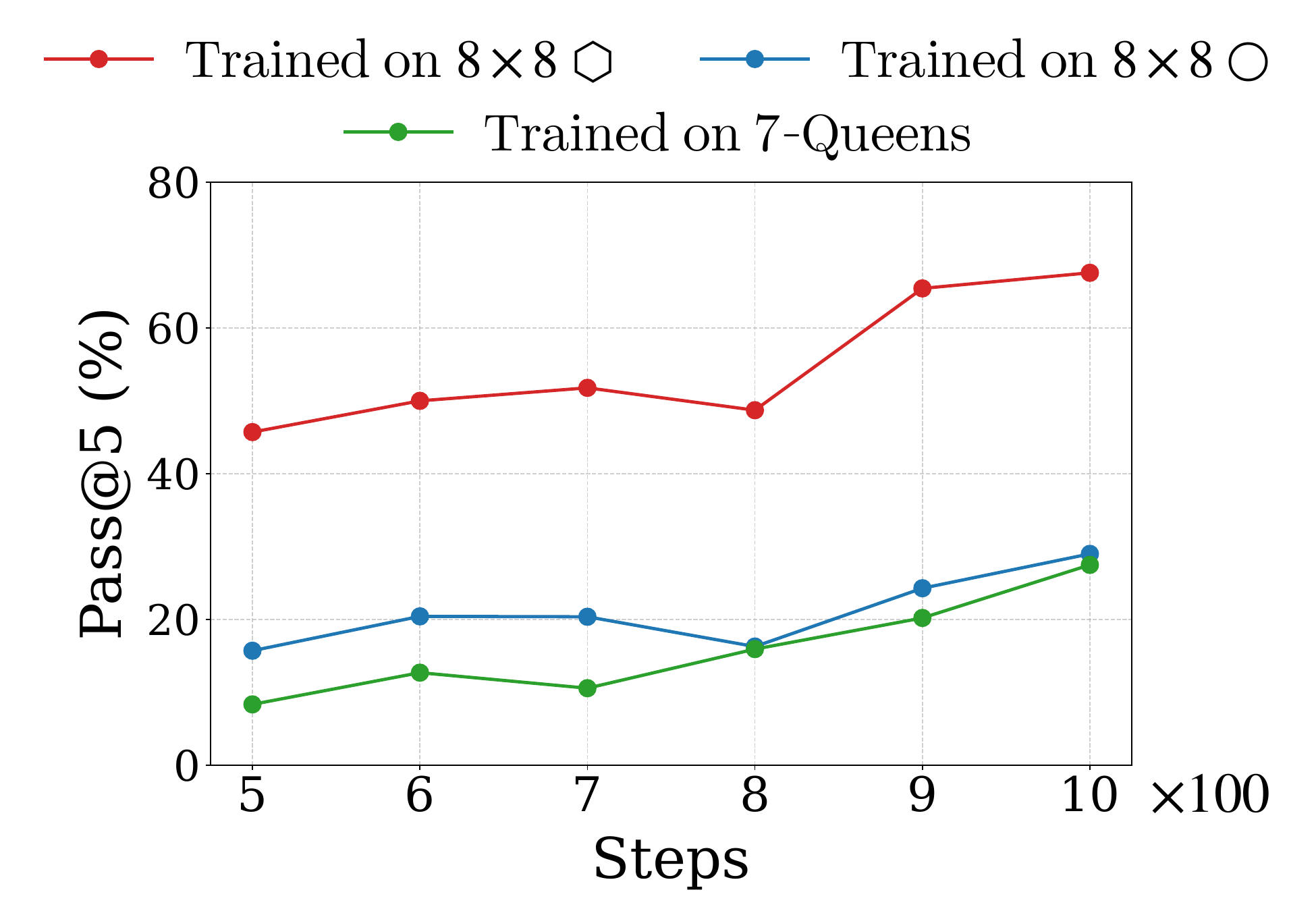} 
     \vspace{-20pt}
    \caption{Compute scaling.
    }
    \label{fig:compute_scaling}
    \vspace{-10pt} 
\end{wrapfigure}

\paragraph{Scaling up training compute.}
We double the training duration from 500 to 1000 steps (equivalent to increasing from 2.5 to 5 epochs) while maintaining a fixed training set of 6400 samples. Overall, scaling up training compute yields consistent improvements except for slight drops on Maze at step 800 and on Queen at step 700. Interestingly, gains are generally marginal over 500--700 steps and become more pronounced from step 700 onward. For example, the performance on $\ishexagon$ improves by 6.1\% over 500--700 steps and by 15.8\% over 700--1000 steps. Given the upward momentum in performance, we hypothesize that extended training will yield further gains.
A more detailed analysis of the interaction between data and compute is provided in Appendix~\ref{app:scaling_more}.

\begin{figure*}[t!]
    \vspace{-20pt}
    \centering
    \includegraphics[width=\textwidth]{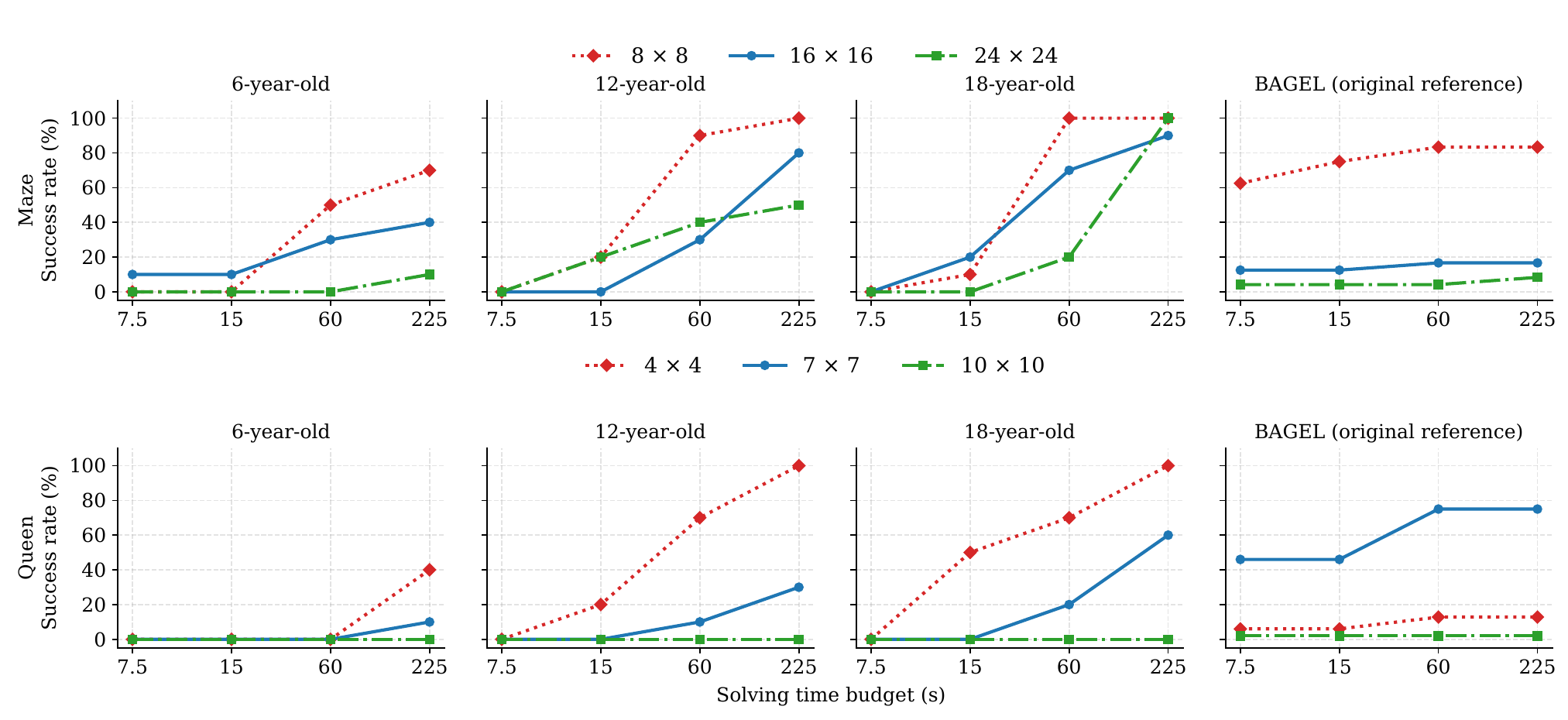} 
    \vspace{-20pt} 
    \caption{Success rates of humans and Bagel under different time budgets. 
    }
    \label{fig:human_study} 
\end{figure*}


\subsection{Human Studies}

We also conducted a comparative study between the model and humans. 

\paragraph{Settings.} We use Bagel as the model representative that is fine-tuned on $8\times8$ $\ishexagon$ mazes and 7-Queens, respectively. For human solvers, we recruited volunteers from three different age groups, representing different stages of cognitive development:
\begin{itemize}[leftmargin=*]
    \item \textbf{6-year-old} represents early childhood, where basic visual planning skills are developed but complex logical planning is still forming.
    \item \textbf{12-year-old} represents the transition to formal operational thoughts, where abstract reasoning and visual planning are largely consolidated.
    \item \textbf{18-year-old} represents the adult baseline for fully mature visual planning.
\end{itemize}

Each age group consists of ten participants; each individual is assigned $\ishexagon$ mazes across three scales ($8\times8$, $16\times16$, and $24\times24$) and Queen puzzles of scales 4, 7, and 10. 
These scales are selected to provide a balanced coverage of difficulty levels within each task, representing easy, moderate, and hard levels respectively.
This configuration yields 30 trials per age group for each task, facilitating controlled analysis across task complexity and cognitive development stage.

We provide participants with unlimited time for mental reasoning prior to drawing their solutions. To align with the model's inference process, participants are required to complete their drawings in a single, continuous attempt—prohibiting erasing, backtracking, or restarts. We record the total latencies for both the reasoning and drawing phases. To ensure a fair comparison, the model is allocated a time budget equivalent to that of human participants, during which it may generate as many candidate solutions as the budget allows.\footnote{A single `drawing' takes the model about 7.5 seconds, averaged over 20 runs.}

\paragraph{The success rate of humans is more positively correlated with the time permitted than that of the model.} Unsurprisingly, with increasing time allowed, human solvers tend to achieve a higher success rate (see Figure~\ref{fig:human_study}), particularly in harder tasks. In contrast, the performance of the model remains relatively flat regardless of the time allowed. Moreover, the elder group demonstrates better leverage of extra time; for example, the 18-year-old group achieves 60\% success rate on 7-Queens within 225 seconds, presumably because their visual planning ability has matured.

\vspace{-5pt}

\begin{wrapfigure}{r}{0.60\linewidth}
    \centering
    \vspace{-25pt} 
    \includegraphics[width=\linewidth]{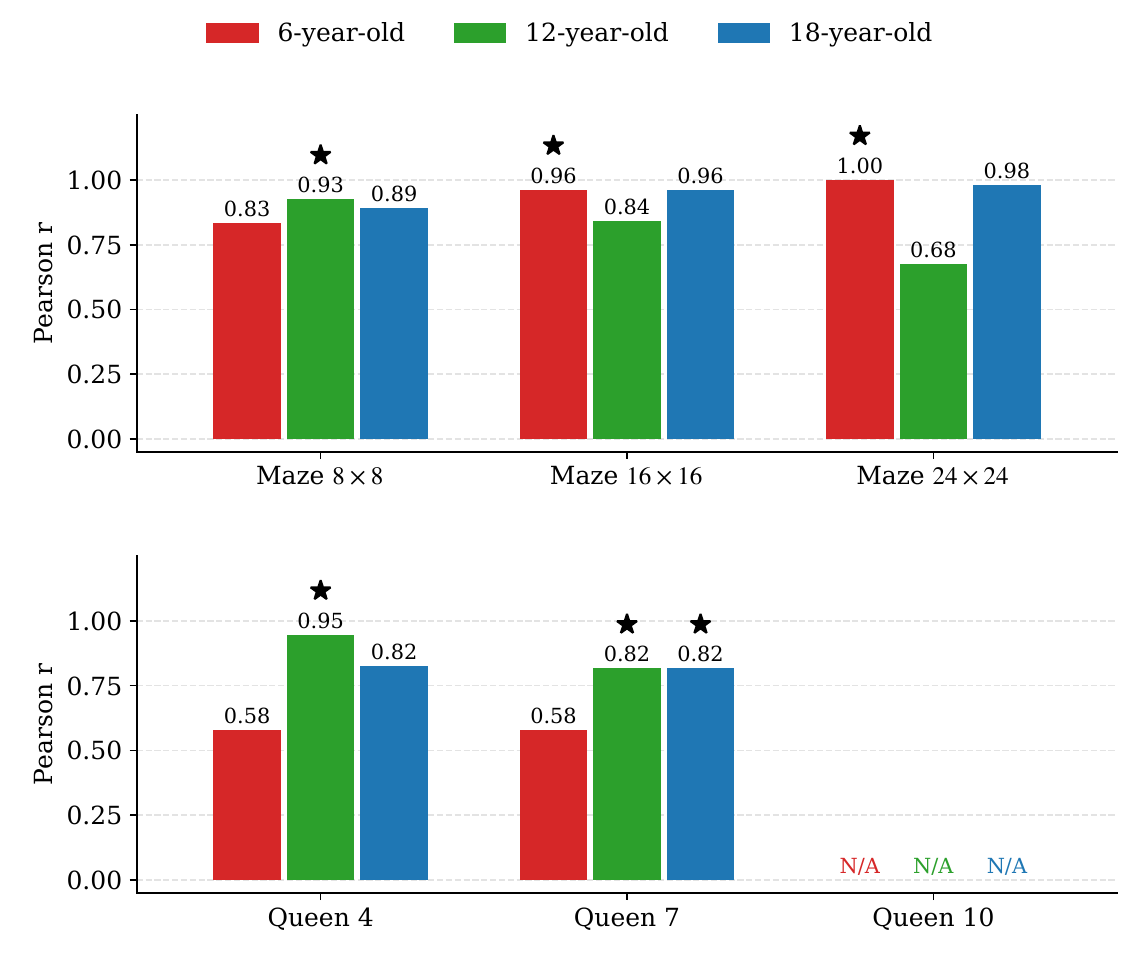}
    \caption{Correlation between model and human groups. Stars mark the highest per task. ``N/A” denotes undefined correlation (zero success rate).
    }
    \label{fig:human_study_correlation}
    \vspace{-10pt} 
\end{wrapfigure}

\paragraph{The visual planning ability of the model resembles that of the 12-year-old on Queen and that of the 6-year-old on Maze.} In general, across tasks of varying difficulty levels, the trend of the model performance does not resemble the same age group (see Figure ~\ref{fig:human_study}). To better understand their relationship, we estimate the Pearson correlation between the model and each human group on each task (see Figure~\ref{fig:human_study_correlation}). On Maze, we observe that the model correlates best with the 6-year-old, and on Queen, it correlates best with the 12-year-old, because that combinatorial planning under the global constraint is generally harder, both for human and the model. 

\section{Related Work}
\textbf{Spatial reasoning.} Spatial reasoning via visual planning requires a deep understanding of topological properties and logical rules. Existing paradigms either rely fully on textual reasoning as a proxy~\citep{ivanitskiy2023maze, dao2025alphamaze} or integrate chain-of-thought prompting into visual reasoning~\citep{wu2025reinforcing, Li2025Imagine, zhang2025reasongen}.
While there has been work exploring fully visual approaches, without relying on textual reasoning~\citep{xu2025letsthink, zhang2025vfscale}, they only consider simple grid-based topologies and use costly step-wise image generation to model sequential planning.
In contrast, we curate a set of spatial reasoning tasks of diverse visual geometries and propose an efficient editing-as-reasoning framework.

\textbf{Image editing models.} 
The goal of image editing is to transform an input image per the given instruction. Existing image editing models generally fall into two main streams: (1) autoregressive models that rely on token-based image representations for causal language-like modeling~\citep{chen2025janus, Chameleon2024}, and (2) diffusion-based models that foster global structural awareness by simultaneously refining the entire image manifold through iterative denoising~\citep{lipman2023flow, deng2025bagel}. Early work learns a standalone editing model~\citep{instructpix2pix2023} while recent research focuses on developing unified multimodal models capable of both image understanding and generation~\citep{Chameleon2024, chen2025janus, deng2025bagel}. We formulate visual spatial reasoning as an editing task and repurpose recent strong editing models for it.

\textbf{Evaluations of image editing models.}  
Evaluations of image editing models assess whether the transformed image aligns with the given instruction. They have been through visual question-answering-based checks~\citep{Antol_2015_ICCV, Goyal_2017_CVPR}, vision-language models based judges~\citep{chen2024mllmasajudge}, and image-text alignment scoring~\citep{Watanabe2023eval, kim2025eval}, but these evaluation paradigms often prioritize semantic fidelity or consistency over logical correctness~\citep{Tong_2024_CVPR, yu2025farvlm}, thus inadequate for tasks that emphasize logical validity.
To address the gap, we curate a set of abstract reasoning tasks devoid of perceptual complexity, accompany it with reliable and rule-based automatic metrics, and evaluate the intrinsic visual planning in image editing models.

%
\vspace{-5pt}
\section{Conclusion}
\vspace{-5pt}
We have proposed \model, an editing-as-reasoning paradigm that reformulates visual planning as a single-step image-editing task. 
To benchmark editing models on visual planning, we develop \bench, a set of abstract visual planning tasks that consist of Maze and Queen, covering two complementary paradigms of visual planning. 
\bench is designed to be devoid of perceptual complexity, enabling a focused study of models' intrinsic visual planning and facilitating reliable and automatic evaluation. 
We empirically find that existing editing models are still limited in abstract visual planning. While supervised fine-tuning on simple tasks yields remarkable improvements, the best fine-tuned model still falls short of the instantaneous, nearly zero-shot reasoning of human solvers.

\section{Acknowledgements}

This work is funded by NSFC-62676236, NSFC-62306046, and NSFC-12574467. We declare no competing interests.

We would like to thank Chenghao Liu and Jian Zhao for their assistance with the experiments and helpful suggestions.

\bibliography{iclr_workshop/iclr2026/iclr2026_conference}

@inproceedings{wu2025reinforcing,
  title     = {Reinforcing {S}patial {R}easoning in {V}ision-{L}anguage {M}odels with {I}nterwoven {T}hinking and {V}isual {D}rawing},
  author    = {Wu, Junfei and Guan, Jian and Feng, Kaituo and Liu, Qiang and Wu, Shu and Wang, Liang and Wu, Wei and Tan, Tieniu},
  booktitle = {Advances in {N}eural {I}nformation {P}rocessing {S}ystems},
  volume    = {38},
  year      = {2025}
}

@misc{codebox_mazes,
  author = {Rob Dawson},
  title = {mazes},
  year = {2021}, 
  publisher = {GitHub},
  journal = {GitHub repository},
  howpublished = {\url{https://github.com/codebox/mazes}},
}

@misc{zhang2025vfscale,
      title={VFScale: Intrinsic Reasoning through Verifier-Free Test-time Scalable Diffusion Model}, 
      author={Tao Zhang and Jia-Shu Pan and Ruiqi Feng and Tailin Wu},
      year={2025},
      eprint={2502.01989},
      archivePrefix={arXiv},
      primaryClass={cs.LG},
      url={https://arxiv.org/abs/2502.01989}, 
}

@misc{wu2025qwenimagetechnicalreport,
      title={Qwen-Image Technical Report}, 
      author={Chenfei Wu and Jiahao Li and Jingren Zhou and Junyang Lin and Kaiyuan Gao and Kun Yan and Sheng-ming Yin and Shuai Bai and Xiao Xu and Yilei Chen and Yuxiang Chen and Zecheng Tang and Zekai Zhang and Zhengyi Wang and An Yang and Bowen Yu and Chen Cheng and Dayiheng Liu and Deqing Li and Hang Zhang and Hao Meng and Hu Wei and Jingyuan Ni and Kai Chen and Kuan Cao and Liang Peng and Lin Qu and Minggang Wu and Peng Wang and Shuting Yu and Tingkun Wen and Wensen Feng and Xiaoxiao Xu and Yi Wang and Yichang Zhang and Yongqiang Zhu and Yujia Wu and Yuxuan Cai and Zenan Liu},
      year={2025},
      eprint={2508.02324},
      archivePrefix={arXiv},
      primaryClass={cs.CV},
      url={https://arxiv.org/abs/2508.02324}, 
}

@InProceedings{Xu_2025_ICCV,
    author    = {Xu, Guowei and Jin, Peng and Wu, Ziang and Li, Hao and Song, Yibing and Sun, Lichao and Yuan, Li},
    title     = {LLaVA-CoT: Let Vision Language Models Reason Step-by-Step},
    booktitle = {Proceedings of the IEEE/CVF International Conference on Computer Vision (ICCV)},
    month     = {October},
    year      = {2025},
    pages     = {2087-2098}
}

@inproceedings{zhang-etal-2025-improve,
    title = "Improve Vision Language Model Chain-of-thought Reasoning",
    author = "Zhang, Ruohong  and
      Zhang, Bowen  and
      Li, Yanghao  and
      Zhang, Haotian  and
      Sun, Zhiqing  and
      Gan, Zhe  and
      Yang, Yinfei  and
      Pang, Ruoming  and
      Yang, Yiming",
    editor = "Che, Wanxiang  and
      Nabende, Joyce  and
      Shutova, Ekaterina  and
      Pilehvar, Mohammad Taher",
    booktitle = "Proceedings of the 63rd Annual Meeting of the Association for Computational Linguistics (Volume 1: Long Papers)",
    month = jul,
    year = "2025",
    address = "Vienna, Austria",
    publisher = "Association for Computational Linguistics",
    url = "https://aclanthology.org/2025.acl-long.82/",
    doi = "10.18653/v1/2025.acl-long.82",
    pages = "1631--1662",
    ISBN = "979-8-89176-251-0",
}

@inproceedings{wang-etal-2025-mathcoder,
    title = "{M}ath{C}oder-{VL}: Bridging Vision and Code for Enhanced Multimodal Mathematical Reasoning",
    author = "Wang, Ke  and
      Pan, Junting  and
      Wei, Linda  and
      Zhou, Aojun  and
      Shi, Weikang  and
      Lu, Zimu  and
      Xiao, Han  and
      Yang, Yunqiao  and
      Ren, Houxing  and
      Zhan, Mingjie  and
      Li, Hongsheng",
    editor = "Che, Wanxiang  and
      Nabende, Joyce  and
      Shutova, Ekaterina  and
      Pilehvar, Mohammad Taher",
    booktitle = "Findings of the Association for Computational Linguistics: ACL 2025",
    month = jul,
    year = "2025",
    address = "Vienna, Austria",
    publisher = "Association for Computational Linguistics",
    url = "https://aclanthology.org/2025.findings-acl.128/",
    doi = "10.18653/v1/2025.findings-acl.128",
    pages = "2505--2534",
    ISBN = "979-8-89176-256-5",
}

@misc{instructpix2pix2023,
      title={InstructPix2Pix: Learning to Follow Image Editing Instructions}, 
      author={Tim Brooks and Aleksander Holynski and Alexei A. Efros},
      year={2023},
      eprint={2211.09800},
      archivePrefix={arXiv},
      primaryClass={cs.CV},
      url={https://arxiv.org/abs/2211.09800}, 
}

@inproceedings{
lipman2023flow,
title={Flow Matching for Generative Modeling},
author={Yaron Lipman and Ricky T. Q. Chen and Heli Ben-Hamu and Maximilian Nickel and Matthew Le},
booktitle={The Eleventh International Conference on Learning Representations },
year={2023},
url={https://openreview.net/forum?id=PqvMRDCJT9t}
}

@article{Chameleon2024,
  author = {Chameleon Team},
  doi = {10.48550/arXiv.2405.09818},
  journal = {arXiv preprint arXiv:2405.09818},
  title = {Chameleon: Mixed-Modal Early-Fusion Foundation Models},
  url = {https://github.com/facebookresearch/chameleon},
  year = {2024}
}

@InProceedings{pmlr-v202-li23q,
  title = 	 {{BLIP}-2: Bootstrapping Language-Image Pre-training with Frozen Image Encoders and Large Language Models},
  author =       {Li, Junnan and Li, Dongxu and Savarese, Silvio and Hoi, Steven},
  booktitle = 	 {Proceedings of the 40th International Conference on Machine Learning},
  pages = 	 {19730--19742},
  year = 	 {2023},
  editor = 	 {Krause, Andreas and Brunskill, Emma and Cho, Kyunghyun and Engelhardt, Barbara and Sabato, Sivan and Scarlett, Jonathan},
  volume = 	 {202},
  series = 	 {Proceedings of Machine Learning Research},
  month = 	 {23--29 Jul},
  publisher =    {PMLR},
  url = 	 {https://proceedings.mlr.press/v202/li23q.html},
}

@InProceedings{Antol_2015_ICCV,
author = {Antol, Stanislaw and Agrawal, Aishwarya and Lu, Jiasen and Mitchell, Margaret and Batra, Dhruv and Zitnick, C. Lawrence and Parikh, Devi},
title = {VQA: Visual Question Answering},
booktitle = {Proceedings of the IEEE International Conference on Computer Vision (ICCV)},
month = {December},
year = {2015}
}

@inproceedings{
chen2024mllmasajudge,
title={{MLLM}-as-a-Judge: Assessing Multimodal {LLM}-as-a-Judge with Vision-Language Benchmark},
author={Dongping Chen and Ruoxi Chen and Shilin Zhang and Yaochen Wang and Yinuo Liu and Huichi Zhou and Qihui Zhang and Yao Wan and Pan Zhou and Lichao Sun},
booktitle={Forty-first International Conference on Machine Learning},
year={2024},
url={https://openreview.net/forum?id=dbFEFHAD79}
}

@InProceedings{Goyal_2017_CVPR,
author = {Goyal, Yash and Khot, Tejas and Summers-Stay, Douglas and Batra, Dhruv and Parikh, Devi},
title = {Making the v in VQA Matter: Elevating the Role of Image Understanding in Visual Question Answering},
booktitle = {Proceedings of the IEEE Conference on Computer Vision and Pattern Recognition (CVPR)},
month = {July},
year = {2017}
}

@InProceedings{Tong_2024_CVPR,
    author    = {Tong, Shengbang and Liu, Zhuang and Zhai, Yuexiang and Ma, Yi and LeCun, Yann and Xie, Saining},
    title     = {Eyes Wide Shut? Exploring the Visual Shortcomings of Multimodal LLMs},
    booktitle = {Proceedings of the IEEE/CVF Conference on Computer Vision and Pattern Recognition (CVPR)},
    month     = {June},
    year      = {2024},
    pages     = {9568-9578}
}

@inproceedings{
xu2025visual,
title={Visual Planning: Let's Think Only with Images},
author={Yi Xu and Chengzu Li and Han Zhou and Xingchen Wan and Caiqi Zhang and Anna Korhonen and Ivan Vuli{\'c}},
booktitle={Workshop on Foundation Models Meet Embodied Agents at CVPR 2025},
year={2025},
url={https://openreview.net/forum?id=ELIt3v3S1J}
}

@misc{yu2025farvlm,
      title={How Far are VLMs from Visual Spatial Intelligence? A Benchmark-Driven Perspective}, 
      author={Songsong Yu and Yuxin Chen and Hao Ju and Lianjie Jia and Fuxi Zhang and Shaofei Huang and Yuhan Wu and Rundi Cui and Binghao Ran and Zaibin Zhang and Zhedong Zheng and Zhipeng Zhang and Yifan Wang and Lin Song and Lijun Wang and Yanwei Li and Ying Shan and Huchuan Lu},
      year={2025},
      eprint={2509.18905},
      archivePrefix={arXiv},
      primaryClass={cs.AI},
      url={https://arxiv.org/abs/2509.18905}, 
}

@Article{Watanabe2023eval,
AUTHOR = {Watanabe, Yuto and Togo, Ren and Maeda, Keisuke and Ogawa, Takahiro and Haseyama, Miki},
TITLE = {Manipulation Direction: Evaluating Text-Guided Image Manipulation Based on Similarity between Changes in Image and Text Modalities},
JOURNAL = {Sensors},
VOLUME = {23},
YEAR = {2023},
NUMBER = {22},
ARTICLE-NUMBER = {9287},
URL = {https://www.mdpi.com/1424-8220/23/22/9287},
PubMedID = {38005673},
ISSN = {1424-8220},
DOI = {10.3390/s23229287}
}

@misc{dao2025alphamaze,
      title={AlphaMaze: Enhancing Large Language Models' Spatial Intelligence via GRPO}, 
      author={Alan Dao and Dinh Bach Vu},
      year={2025},
      eprint={2502.14669},
      archivePrefix={arXiv},
      primaryClass={cs.CL},
      url={https://arxiv.org/abs/2502.14669}, 
}

@inproceedings{kim2025eval,
  author={Kim, Yoonjeon and Ryu, Soohyun and Jung, Yeonsung and Lee, Hyunkoo and Kim, Joowon and Yang, June Yong and Hwang, Jaeryong and Yang, Eunho},
  booktitle={2025 IEEE/CVF Conference on Computer Vision and Pattern Recognition (CVPR)}, 
  title={Preserve or Modify? Context-Aware Evaluation for Balancing Preservation and Modification in Text-Guided Image Editing}, 
  year={2025},
  volume={},
  number={},
  pages={23474-23483},
  doi={10.1109/CVPR52734.2025.02186}}

@misc{ivanitskiy2023maze,
      title={Structured World Representations in Maze-Solving Transformers}, 
      author={Michael Igorevich Ivanitskiy and Alex F. Spies and Tilman Räuker and Guillaume Corlouer and Chris Mathwin and Lucia Quirke and Can Rager and Rusheb Shah and Dan Valentine and Cecilia Diniz Behn and Katsumi Inoue and Samy Wu Fung},
      year={2023},
      eprint={2312.02566},
      archivePrefix={arXiv},
      primaryClass={cs.LG},
      url={https://arxiv.org/abs/2312.02566}, 
}

@misc{xu2025letsthink,
      title={Visual Planning: Let's Think Only with Images}, 
      author={Yi Xu and Chengzu Li and Han Zhou and Xingchen Wan and Caiqi Zhang and Anna Korhonen and Ivan Vulić},
      year={2025},
      eprint={2505.11409},
      archivePrefix={arXiv},
      primaryClass={cs.LG},
      url={https://arxiv.org/abs/2505.11409}, 
}

@InProceedings{Li2025Imagine,
  title = 	 {Imagine While Reasoning in Space: Multimodal Visualization-of-Thought},
  author =       {Li, Chengzu and Wu, Wenshan and Zhang, Huanyu and Xia, Yan and Mao, Shaoguang and Dong, Li and Vuli\'{c}, Ivan and Wei, Furu},
  booktitle = 	 {Proceedings of the 42nd International Conference on Machine Learning},
  pages = 	 {36340--36364},
  year = 	 {2025},
  editor = 	 {Singh, Aarti and Fazel, Maryam and Hsu, Daniel and Lacoste-Julien, Simon and Berkenkamp, Felix and Maharaj, Tegan and Wagstaff, Kiri and Zhu, Jerry},
  volume = 	 {267},
  series = 	 {Proceedings of Machine Learning Research},
  month = 	 {13--19 Jul},
  publisher =    {PMLR},
  url = 	 {https://proceedings.mlr.press/v267/li25cz.html}
}

@misc{zhang2025reasongen,
      title={ReasonGen-R1: CoT for Autoregressive Image generation models through SFT and RL}, 
      author={Yu Zhang and Yunqi Li and Yifan Yang and Rui Wang and Yuqing Yang and Dai Qi and Jianmin Bao and Dongdong Chen and Chong Luo and Lili Qiu},
      year={2025},
      eprint={2505.24875},
      archivePrefix={arXiv},
      primaryClass={cs.CV},
      url={https://arxiv.org/abs/2505.24875}, 
}

@inproceedings{yang2022empirical,
  title={An empirical study of gpt-3 for few-shot knowledge-based vqa},
  author={Yang, Zhengyuan and Gan, Zhe and Wang, Jianfeng and Hu, Xiaowei and Lu, Yumao and Liu, Zicheng and Wang, Lijuan},
  booktitle={Proceedings of the AAAI conference on artificial intelligence},
  volume={36},
  pages={3081--3089},
  year={2022}
}

@article{DBLP:journals/corr/abs-2303-04671,
  author       = {Chenfei Wu and
                  Shengming Yin and
                  Weizhen Qi and
                  Xiaodong Wang and
                  Zecheng Tang and
                  Nan Duan},
  title        = {Visual ChatGPT: Talking, Drawing and Editing with Visual Foundation
                  Models},
  journal      = {CoRR},
  volume       = {abs/2303.04671},
  year         = {2023},
  url          = {https://doi.org/10.48550/arXiv.2303.04671},
  doi          = {10.48550/ARXIV.2303.04671},
  eprinttype    = {arXiv},
  eprint       = {2303.04671},
  bibsource    = {dblp computer science bibliography, https://dblp.org}
}

@article{chen2025janus,
  title={Janus-Pro: Unified Multimodal Understanding and Generation with Data and Model Scaling},
  author={Chen, Xiaokang and Wu, Zhiyu and Liu, Xingchao and Pan, Zizheng and Liu, Wen and Xie, Zhenda and Yu, Xingkai and Ruan, Chong},
  journal={arXiv preprint arXiv:2501.17811},
  year={2025}
}

@misc{videozeroshot,
      title={Video models are zero-shot learners and reasoners}, 
      author={Thaddäus Wiedemer and Yuxuan Li and Paul Vicol and Shixiang Shane Gu and Nick Matarese and Kevin Swersky and Been Kim and Priyank Jaini and Robert Geirhos},
      year={2025},
      eprint={2509.20328},
      archivePrefix={arXiv},
      primaryClass={cs.LG},
      url={https://arxiv.org/abs/2509.20328}, 
}

@inproceedings{FID,
 author = {Heusel, Martin and Ramsauer, Hubert and Unterthiner, Thomas and Nessler, Bernhard and Hochreiter, Sepp},
 booktitle = {Advances in Neural Information Processing Systems},
 editor = {I. Guyon and U. Von Luxburg and S. Bengio and H. Wallach and R. Fergus and S. Vishwanathan and R. Garnett},
 pages = {},
 publisher = {Curran Associates, Inc.},
 title = {GANs Trained by a Two Time-Scale Update Rule Converge to a Local Nash Equilibrium},
 url = {https://proceedings.neurips.cc/paper_files/paper/2017/file/8a1d694707eb0fefe65871369074926d-Paper.pdf},
 volume = {30},
 year = {2017}
}

@InProceedings{LPIPS,
author = {Zhang, Richard and Isola, Phillip and Efros, Alexei A. and Shechtman, Eli and Wang, Oliver},
title = {The Unreasonable Effectiveness of Deep Features as a Perceptual Metric},
booktitle = {Proceedings of the IEEE Conference on Computer Vision and Pattern Recognition (CVPR)},
month = {June},
year = {2018}
}

@article{deng2025bagel,
  title   = {Emerging Properties in Unified Multimodal Pretraining},
  author  = {Deng, Chaorui and Zhu, Deyao and Li, Kunchang and Gou, Chenhui and Li, Feng and Wang, Zeyu and Zhong, Shu and Yu, Weihao and Nie, Xiaonan and Song, Ziang and Shi, Guang and Fan, Haoqi},
  journal = {arXiv preprint arXiv:2505.14683},
  year    = {2025}
}

@misc{gptimage1,
  author       = {OpenAI},
  title        = {GPT Image 1: State-of-the-art image generation model},
  howpublished = {Web page},
  url = {https://platform.openai.com/docs/models/gpt-image-1},
  year         = {2025}
}

@misc{gemini3pro_image_2025,
  title        = {Gemini 3 Pro Image (Nano Banana Pro)},
  howpublished = {Web page},
  author    = {Google DeepMind},
  year         = {2025},
  url= {https://deepmind.google/models/gemini-image/pro/}
}

@misc{seedream2025seedream40nextgenerationmultimodal,
      title={Seedream 4.0: Toward Next-generation Multimodal Image Generation}, 
      author={Team Seedream and : and Yunpeng Chen and Yu Gao and Lixue Gong and Meng Guo and Qiushan Guo and Zhiyao Guo and Xiaoxia Hou and Weilin Huang and Yixuan Huang and Xiaowen Jian and Huafeng Kuang and Zhichao Lai and Fanshi Li and Liang Li and Xiaochen Lian and Chao Liao and Liyang Liu and Wei Liu and Yanzuo Lu and Zhengxiong Luo and Tongtong Ou and Guang Shi and Yichun Shi and Shiqi Sun and Yu Tian and Zhi Tian and Peng Wang and Rui Wang and Xun Wang and Ye Wang and Guofeng Wu and Jie Wu and Wenxu Wu and Yonghui Wu and Xin Xia and Xuefeng Xiao and Shuang Xu and Xin Yan and Ceyuan Yang and Jianchao Yang and Zhonghua Zhai and Chenlin Zhang and Heng Zhang and Qi Zhang and Xinyu Zhang and Yuwei Zhang and Shijia Zhao and Wenliang Zhao and Wenjia Zhu},
      year={2025},
      eprint={2509.20427},
      archivePrefix={arXiv},
      primaryClass={cs.CV},
      url={https://arxiv.org/abs/2509.20427}, 
}

@misc{labs2025flux1kontextflowmatching,
      title={FLUX.1 Kontext: Flow Matching for In-Context Image Generation and Editing in Latent Space},
      author={Black Forest Labs and Stephen Batifol and Andreas Blattmann and Frederic Boesel and Saksham Consul and Cyril Diagne and Tim Dockhorn and Jack English and Zion English and Patrick Esser and Sumith Kulal and Kyle Lacey and Yam Levi and Cheng Li and Dominik Lorenz and Jonas Müller and Dustin Podell and Robin Rombach and Harry Saini and Axel Sauer and Luke Smith},
      year={2025},
      eprint={2506.15742},
      archivePrefix={arXiv},
      primaryClass={cs.GR},
      url={https://arxiv.org/abs/2506.15742},
}

@article{unifiedreward,
  title={Unified reward model for multimodal understanding and generation},
  author={Wang, Yibin and Zang, Yuhang and Li, Hao and Jin, Cheng and Wang, Jiaqi},
  journal={arXiv preprint arXiv:2503.05236},
  year={2025}
}
\bibliographystyle{iclr2026_conference}
                                                              
%
\newpage
\appendix
\onecolumn

\section{Complete Prompts for Maze and Queen Tasks}
\label{app:instruction}

We provide the complete prompts used for the Maze and Queen tasks. 
For each task, we include the prompts with and without Chain-of-Thought (CoT) as follows:

\vspace{0.5em}
\subsection{Prompts without Chain-of-Thought}

\paragraph{Maze task (without CoT)} requires generating a valid path from the entrance to the exit while strictly following geometric constraints.

\begin{definitionbox2}
Add the blue solution path for the maze, connect start point (solid red circle) to end point (red 'X' mark). Ensure all original maze elements (walls, points, etc.) remain unchanged—only add the path.
\end{definitionbox2}

\paragraph{Queen task (without CoT)} requires placing all queens such that no conflicts occur across rows, columns, and different color regions.

\begin{definitionbox2}
Generate the solved board by placing one queen (represented by a solid black circle in the center of a grid cell) in each row, column, and colored region while ensuring queens do not touch in 8-neighborhood.
\end{definitionbox2}


\subsection{Prompts with Chain-of-Thought (CoT)}

The CoT-augmented prompts explicitly encourage the model to perform intermediate reasoning before producing the final output. 

\paragraph{Maze Task (CoT)} augments the instruction with an additional prompt, as shown below:

\begin{definitionbox2}
Add the blue solution path for the maze, connect start point (solid red circle) to end point (red 'X' mark). Ensure all original maze elements (walls, points, etc.) remain unchanged—only add the path.

You should first think about the planning process in the mind. The planning process must be enclosed within \texttt{<think>} and \texttt{</think>} tags.
\end{definitionbox2}

\paragraph{Queen Task (CoT)} uses a similar process, as shown below:

\begin{definitionbox2}
Generate the solved board by placing one queen (represented by a solid black circle in the center of a grid cell) in each row, column, and colored region while ensuring queens do not touch in 8-neighborhood.

You should first think about the planning process in the mind. The planning process must be enclosed within \texttt{<think>} and \texttt{</think>} tags.
\end{definitionbox2}

\vspace{2em}

For models that do not natively support joint text-and-image generation (e.g., Janus-Pro), we adopt a two-stage inference prompts. 
Prompts for these models consists of two stages: text generation and image generation.
This formulation is shared across both the Maze and Queen tasks. We illustrate the prompt using the Maze task as an example; the same formulation applies to the Queen task. 

\paragraph{Prompt for text generation} requires model to output text CoT, shown as follows:

\vspace{0.5em}

\begin{definitionbox2}
Add the blue solution path for the maze, connect start point (solid red circle) to end point (red 'X' mark). Ensure all original maze elements (walls, points, etc.) remain unchanged—only add the path.

You should first think about the planning process in the mind. The planning process is enclosed within \texttt{<think>} \texttt{</think>} tags. 
\end{definitionbox2}

\paragraph{Prompt for image generation} requires model to output final image only. The ellipsis denotes the model's reasoning in text generation. The prompt is shown as follows:

\vspace{0.5em}

\begin{definitionbox2}
Add the blue solution path for the maze, connect start point (solid red circle) to end point (red 'X' mark). Ensure all original maze elements (walls, points, etc.) remain unchanged—only add the path.

\texttt{<think>}......
\texttt{</think>}

According to your thinking process, output the image only.

\end{definitionbox2}

\section{Scaling Up Training Data on Cross-Domain Performance}
\label{app:data_scaling_ood}

\begin{wrapfigure}{r}{0.5\linewidth}
    \centering
    \vspace{-12pt} 
    \includegraphics[width=\linewidth]{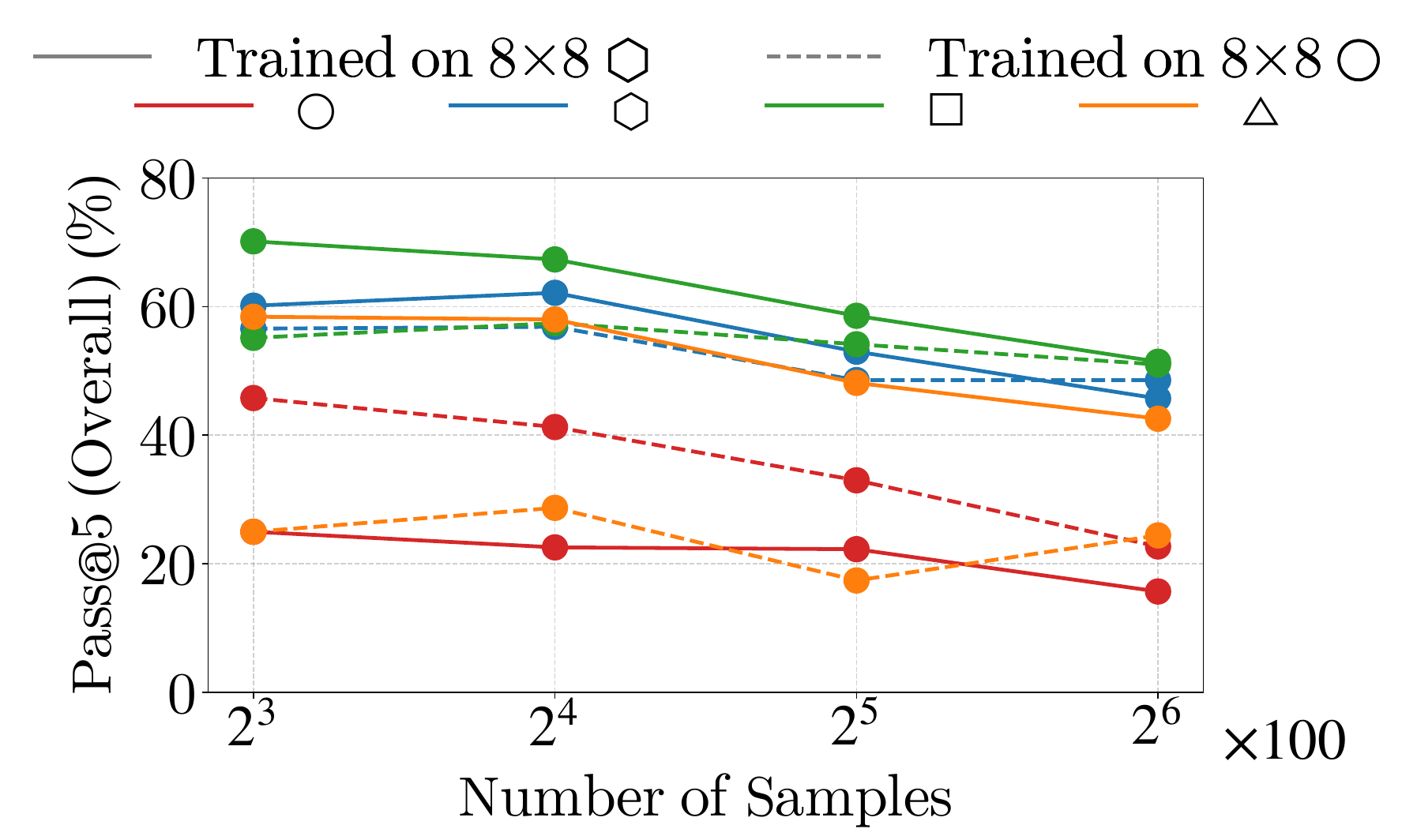} 
    \caption{Data scaling on cross-domain performance.}
    \label{fig:data_scaling_ood}
    \vspace{-10pt} 
\end{wrapfigure}

We further investigate how scaling the training data affects cross-domain performance, where models are trained on a single geometry and evaluated across different geometries. We train models on $8\times8$ $\ishexagon$ mazes and $8\times8$ $\iscircle$ mazes with fixed steps (500), and evaluate cross-domain performance on all geometry types across all scales from $3\times3$ to $16\times16$.

As shown in Figure~\ref{fig:data_scaling_ood}, the topology of the training geometry plays a critical role in determining transferability. Models trained on $\ishexagon$ mazes (solid line) exhibit robust performance across all tested shapes, whereas those trained on $\iscircle$ mazes (dotted line) show weaker transferability. This is primarily because the $\ishexagon$ mazes allow the model to learn stable, translation-invariant navigation strategies, compared to $\iscircle$ mazes with arbitrary action spaces. Interestingly, all models have best performance in $\issquare$ mazes, which is possibly because its trade-off between action space and topological constraints.

Notably, the degradation at larger training data indicates a tendency toward geometry-specific overfitting: as the training distribution becomes denser, the model increasingly specializes to the source geometry, reducing its ability to generalize to structurally different domains.

\section{Extended Analysis of Data–Compute Scaling}
\label{app:scaling_more}

The joint data--compute experiments use $8\times8$ Hexagon mazes, $8\times8$ Circle mazes, and 7-Queens, with $N\in\{800,1600,3200,6400\}$ and checkpoints from 500 to 1,000 steps. Evaluation covers all test scales of the corresponding task and geometry.

\begin{figure}[htbp]  
    \centering
    \includegraphics[width=\linewidth]{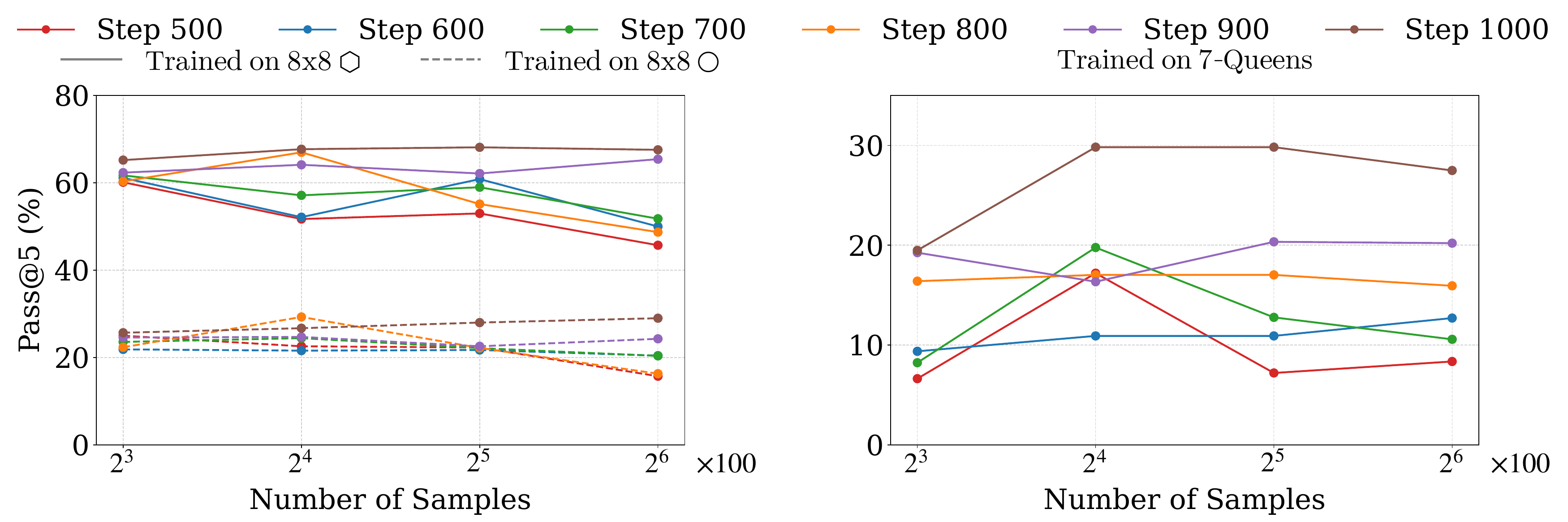} 
    \caption{Joint Scaling of Data and Compute.}
    \label{fig:data_compute_scaling}
    
\end{figure}
Figure~\ref{fig:data_compute_scaling} compares training budgets across all four data sizes. At 1,000 steps, 3,200 samples reach 68.14\% on Hexagon Maze, compared with 67.71\% for 1,600 samples. On Circle Maze, 6,400 samples achieve 29.00\%, compared with 26.71\% for 1,600 samples. The apparent plateau beyond 1,600 samples therefore depends on the optimization budget rather than forming a fixed ceiling. For Queen, 3,200 samples match the 1,600-sample result of 29.83\%, while 6,400 samples reach 27.49\% within the 1,000-step budget. Jointly increasing data and compute improves Maze performance beyond the smaller-data settings; the scaling behavior of Queen remains unresolved within the tested range.

With 6,400 training samples, increasing training from 500 to 1,000 steps improves \textsc{Pass@5} by 21.86 percentage points on Hexagon Maze and 13.29 points on Circle Maze. Larger datasets are thus substantially under-optimized at the smaller compute budget.

\section{Additional Qualitative Studies}
\label{app:qualitative_more}
\begin{figure*}[t!]\small
\centering
    \begin{subfigure}[t]{0.15\textwidth}
        \centering
        \includegraphics[width=\linewidth]{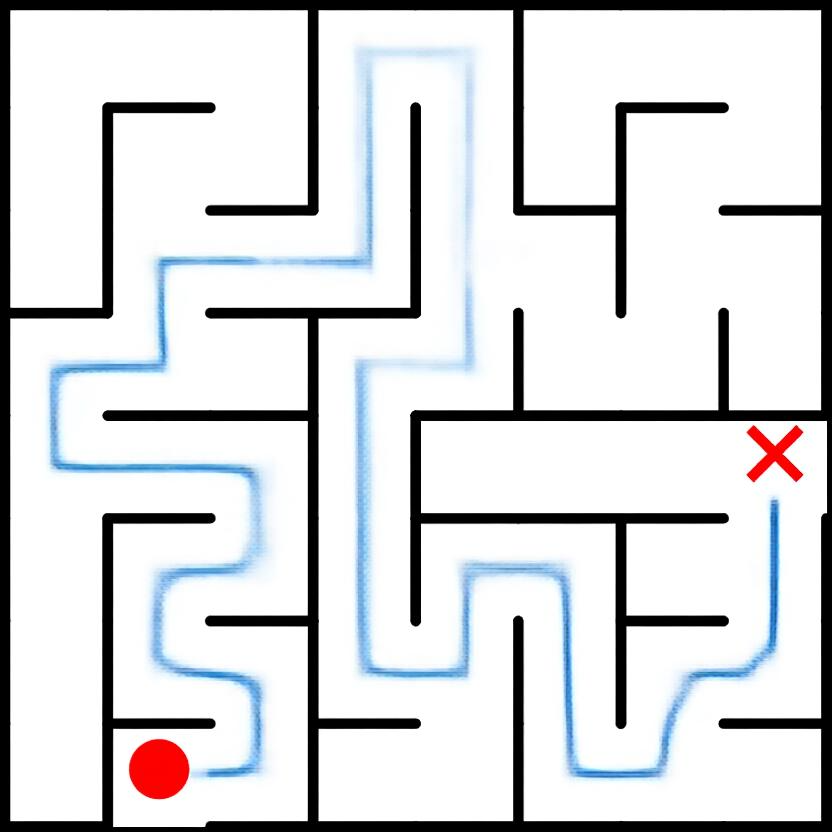}
    \end{subfigure}
    \hfill
    \begin{subfigure}[t]{0.15\textwidth}
        \centering
        \includegraphics[width=\linewidth]{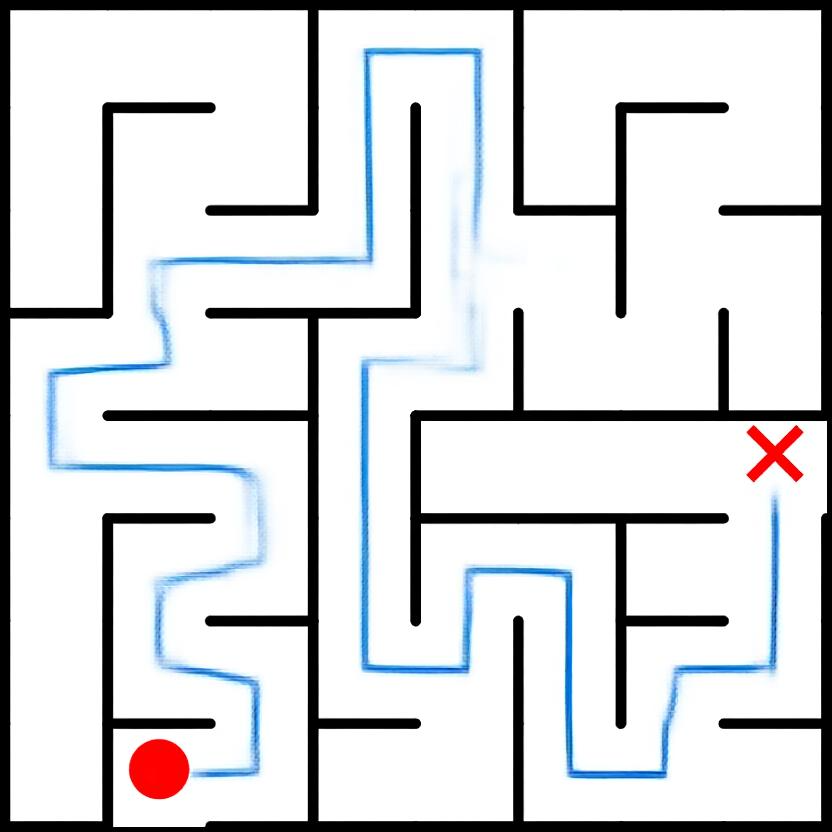}
    \end{subfigure}
    \hfill
    \begin{subfigure}[t]{0.15\textwidth}
        \centering
        \includegraphics[width=\linewidth]{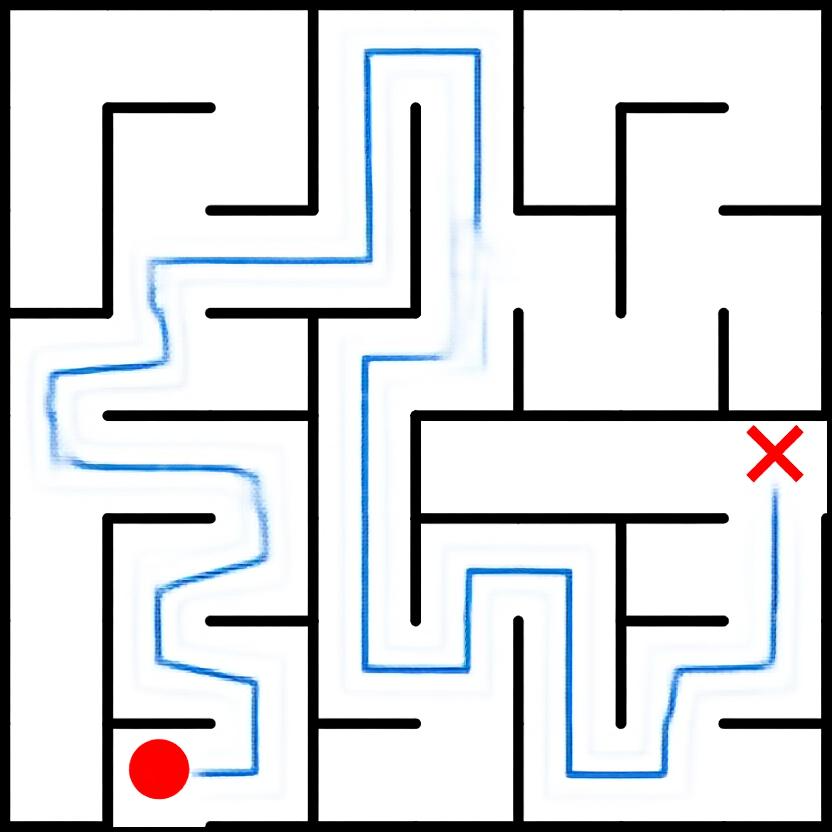}
    \end{subfigure}
    \hfill
    \begin{subfigure}[t]{0.15\textwidth}
        \centering
        \includegraphics[width=\linewidth]{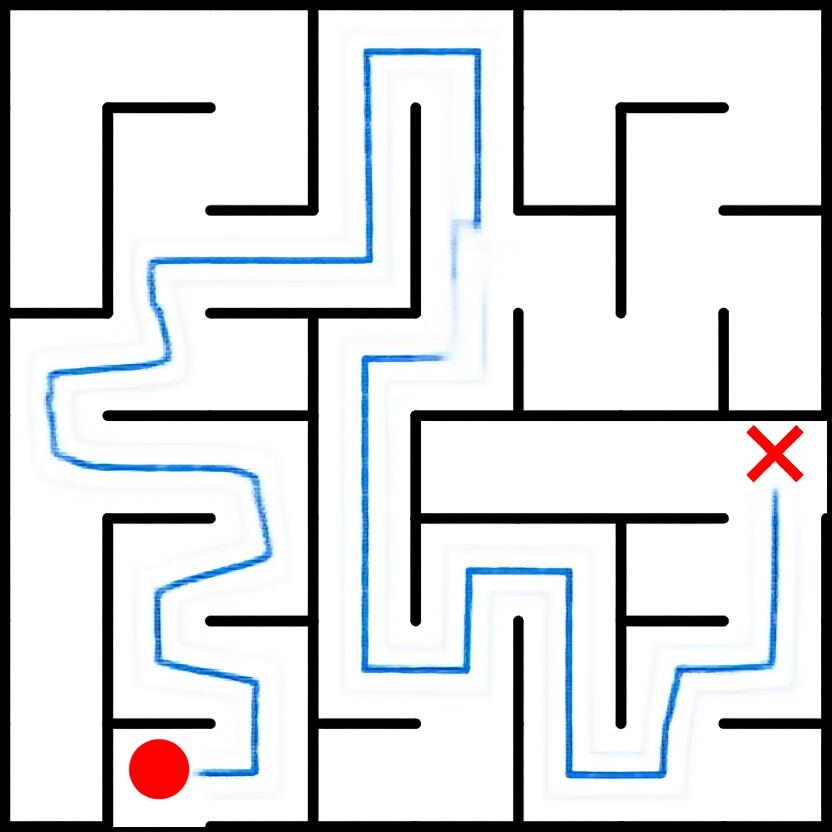}
    \end{subfigure}
    \hfill
    \begin{subfigure}[t]{0.15\textwidth}
        \centering
        \includegraphics[width=\linewidth]{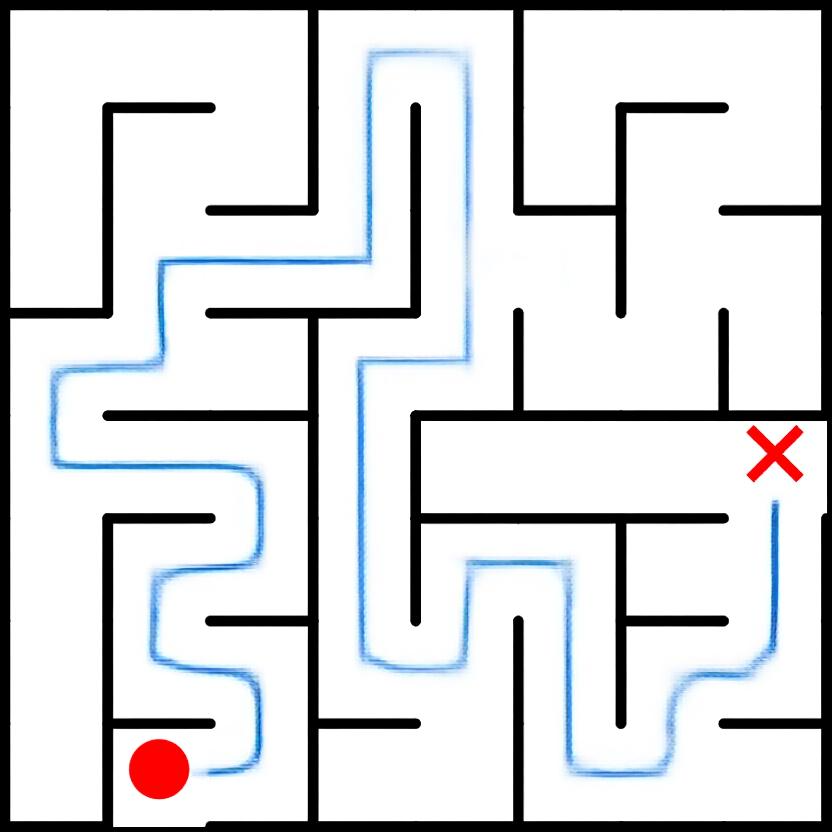}
    \end{subfigure}
    \hfill
    \begin{subfigure}[t]{0.15\textwidth}
        \centering
        \includegraphics[width=\linewidth]{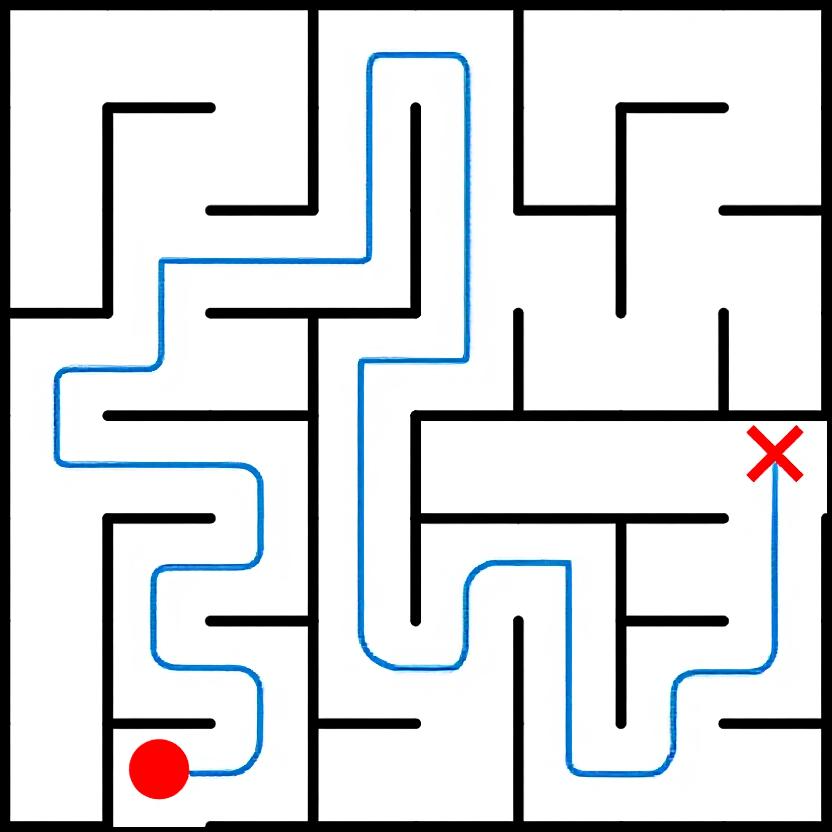}
    \end{subfigure}

    \par\smallskip 

    \begin{subfigure}[t]{0.15\textwidth}
        \centering
        \includegraphics[width=\linewidth]{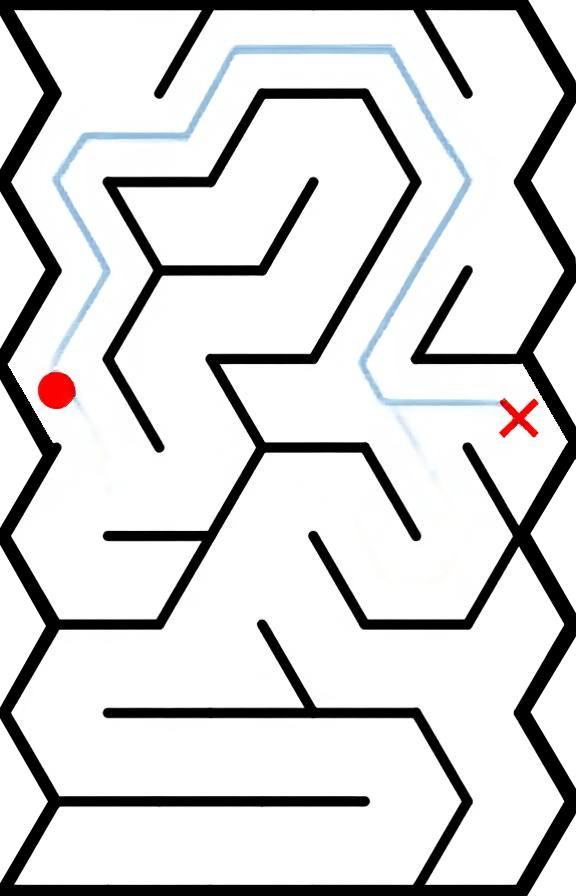}
    \end{subfigure}
    \hfill
    \begin{subfigure}[t]{0.15\textwidth}
        \centering
        \includegraphics[width=\linewidth]{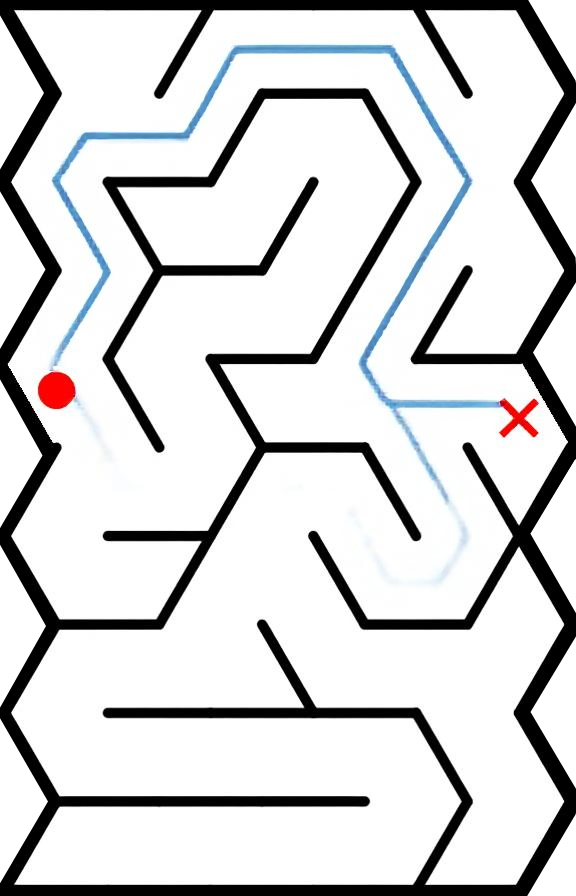}
    \end{subfigure}
    \hfill
    \begin{subfigure}[t]{0.15\textwidth}
        \centering
        \includegraphics[width=\linewidth]{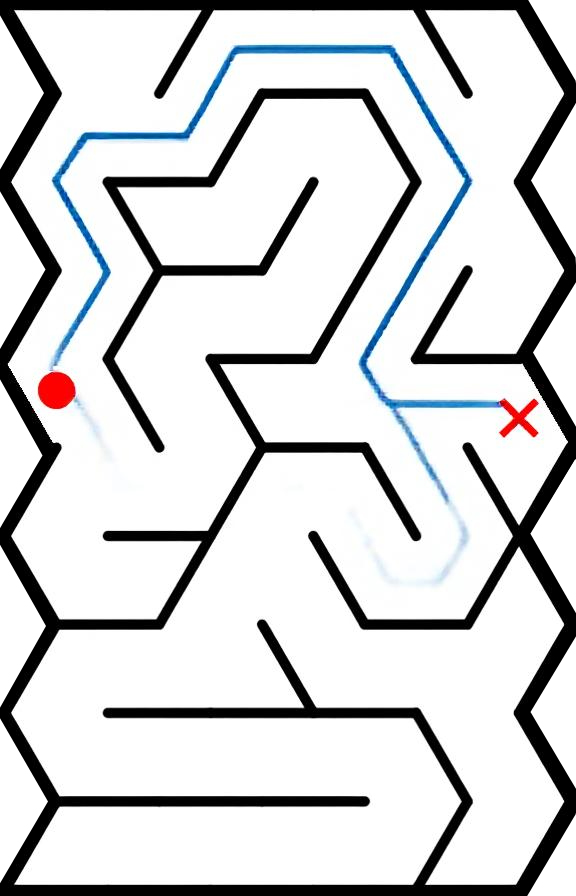}
    \end{subfigure}
    \hfill
    \begin{subfigure}[t]{0.15\textwidth}
        \centering
        \includegraphics[width=\linewidth]{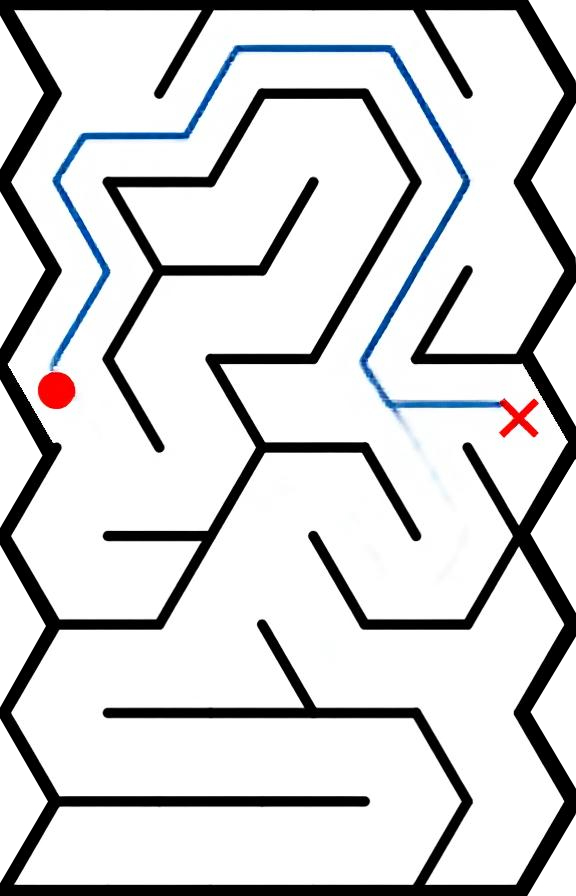}
    \end{subfigure}
    \hfill
    \begin{subfigure}[t]{0.15\textwidth}
        \centering
        \includegraphics[width=\linewidth]{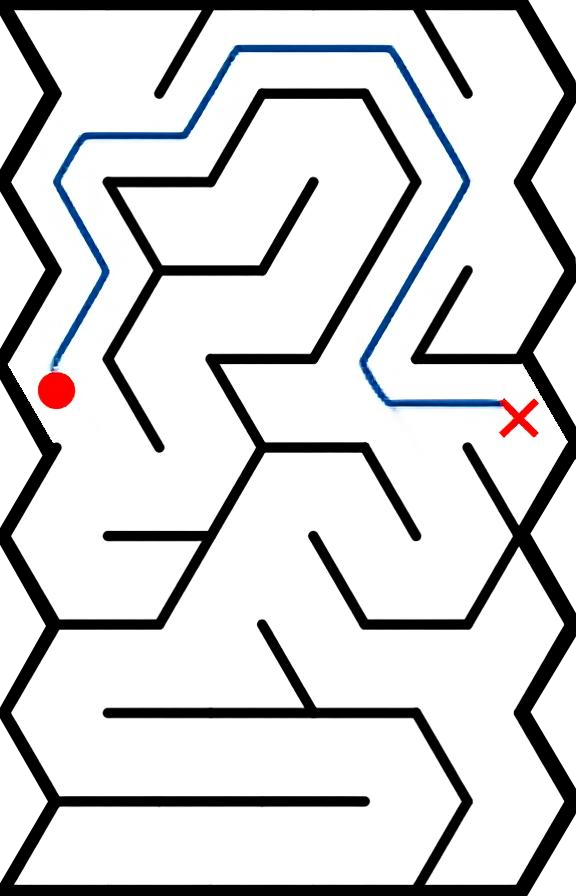}
    \end{subfigure}
    \hfill
    \begin{subfigure}[t]{0.15\textwidth}
        \centering
        \includegraphics[width=\linewidth]{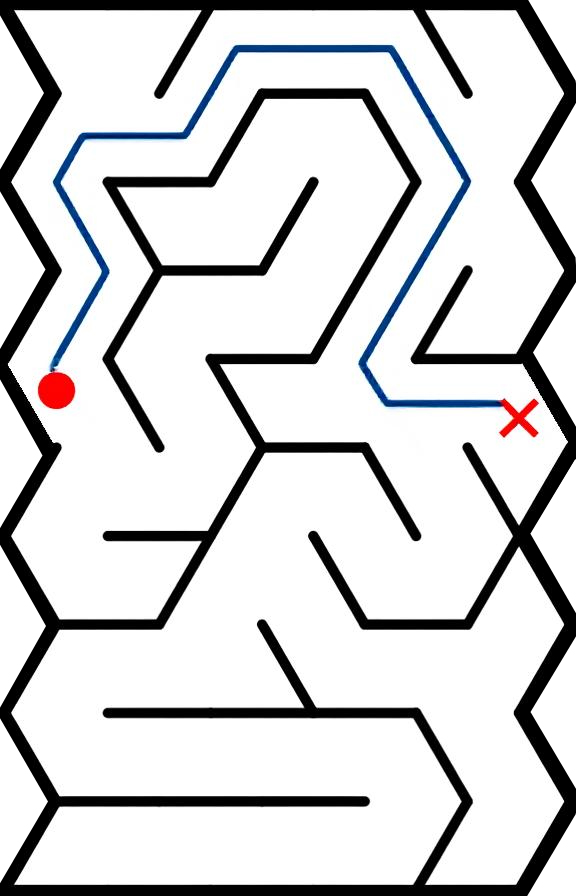}
    \end{subfigure}
        

    \caption{More example solutions from different denoising steps ($t$) of a fine-tuned Bagel on Maze ($\issquare$ and $\istriangle$) task. 
    }
    \label{fig:diffusion_cot_appendix}
    \vspace{-12pt} 
\end{figure*}
To complement the qualitative analysis in the main text, we provide additional examples of the planning process for the Maze task under different geometries, including square and triangle mazes (Figure~\ref{fig:diffusion_cot_appendix}). 

Across these geometries, we observe a consistent coarse-to-fine planning pattern. In early denoising steps, the model tends to form a tentative global trajectory connecting the start and end points, often with low confidence and visible ambiguity. As the denoising process progresses, this trajectory is gradually refined, with incorrect or inconsistent subpaths being revised or removed.

\section{Error Analysis}
\label{app:fatal_case_more}
\begin{figure*}[t!]
    \centering
    \small
    \captionsetup[subfigure]{labelformat=empty} 

    \newcommand{\imgheight}{2.4cm}

    \begin{minipage}[t]{0.48\textwidth}
        \centering
        \begin{subfigure}[b]{0.32\linewidth}
            \centering
            \includegraphics[height=\imgheight, width=\linewidth, keepaspectratio]{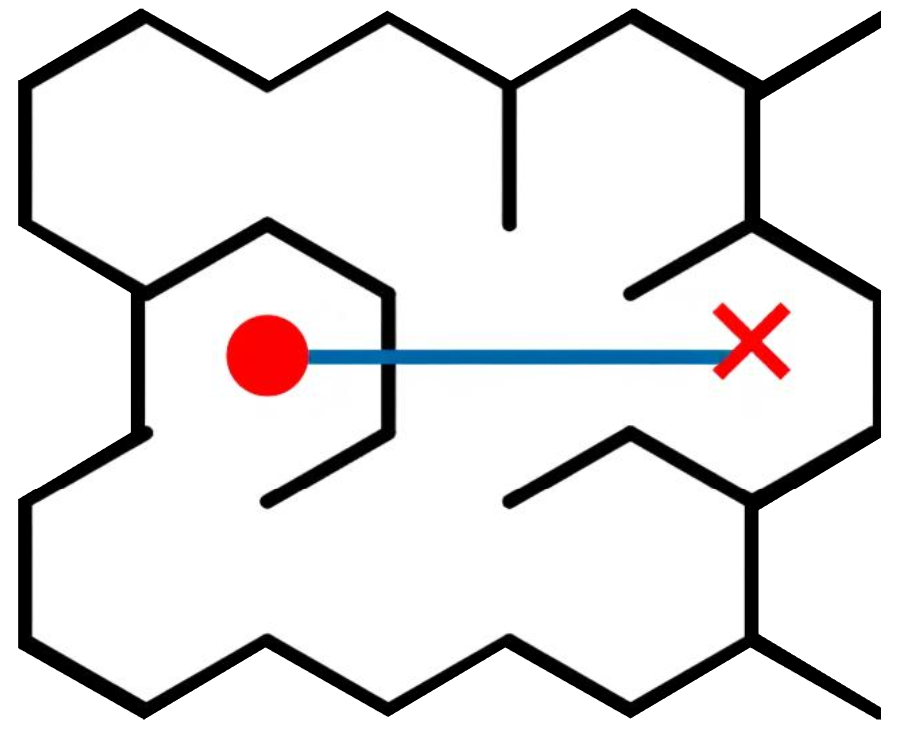}
        \end{subfigure}\hfill
        \begin{subfigure}[b]{0.32\linewidth}
            \centering
            \includegraphics[height=\imgheight, width=\linewidth, keepaspectratio]{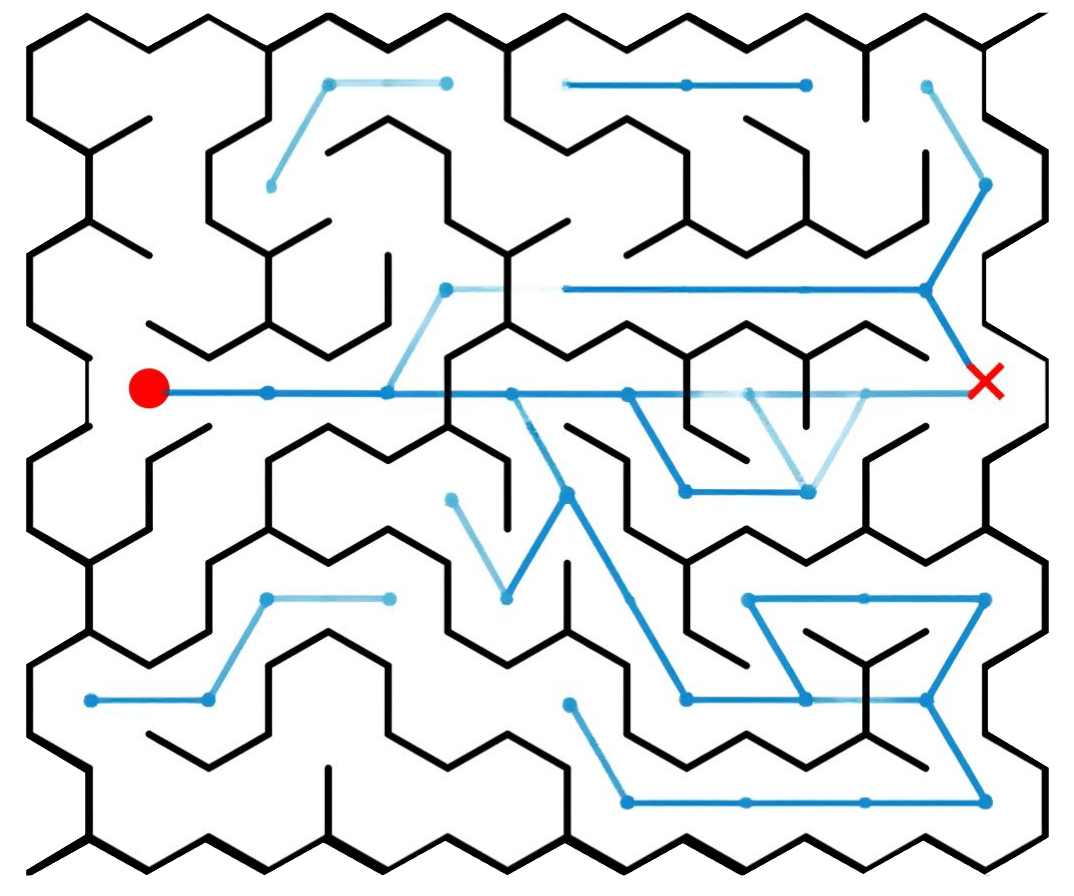}
        \end{subfigure}\hfill
        \begin{subfigure}[b]{0.32\linewidth}
            \centering
            \includegraphics[height=\imgheight, width=\linewidth, keepaspectratio]{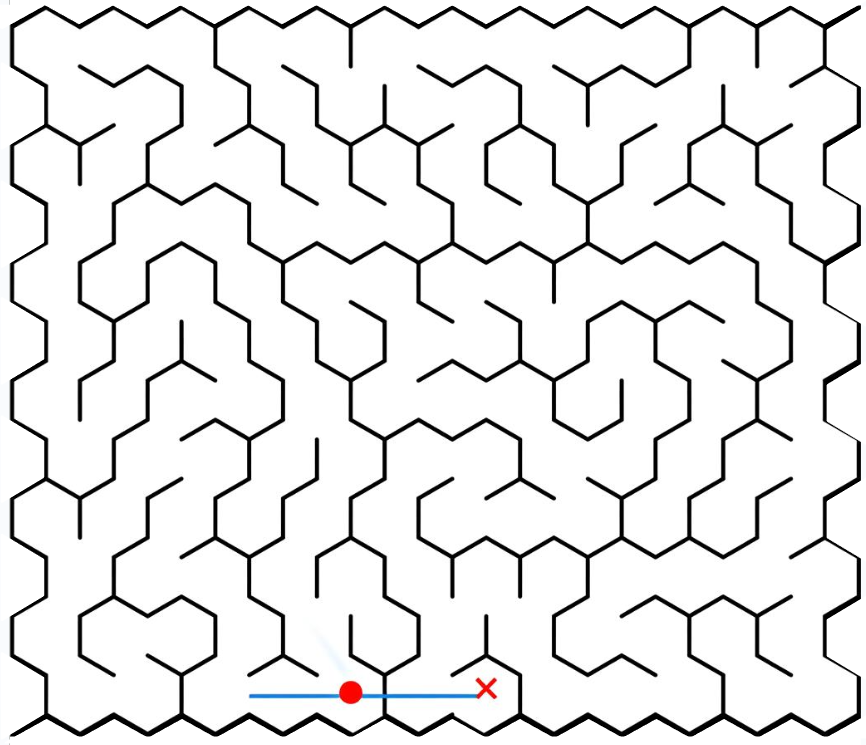}
        \end{subfigure}

        \vspace{0.5em}

        \begin{subfigure}[b]{0.32\linewidth}
            \centering
            \includegraphics[height=\imgheight, width=\linewidth, keepaspectratio]{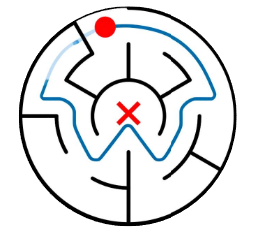}
        \end{subfigure}\hfill
        \begin{subfigure}[b]{0.32\linewidth}
            \centering
            \includegraphics[height=\imgheight, width=\linewidth, keepaspectratio]{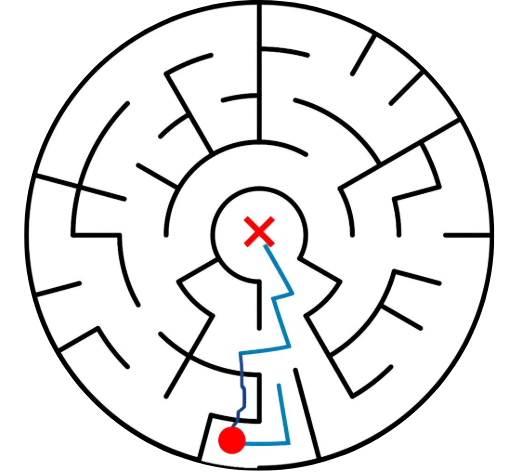}
        \end{subfigure}\hfill
        \begin{subfigure}[b]{0.32\linewidth}
            \centering
            \includegraphics[height=\imgheight, width=\linewidth, keepaspectratio]{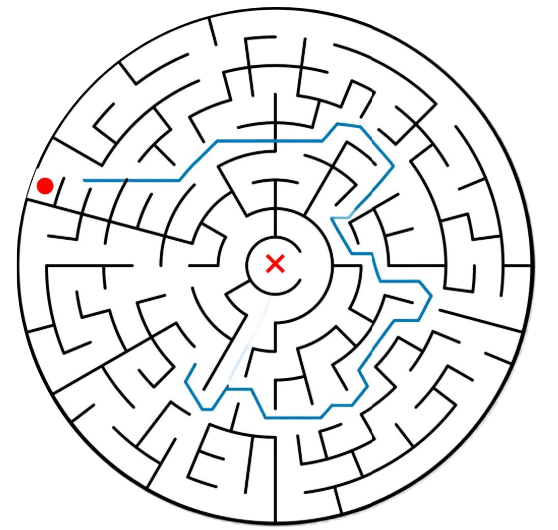}
        \end{subfigure}

        \vspace{0.5em}

        \begin{subfigure}[b]{0.32\linewidth}
            \centering
            \includegraphics[height=\imgheight, width=\linewidth, keepaspectratio]{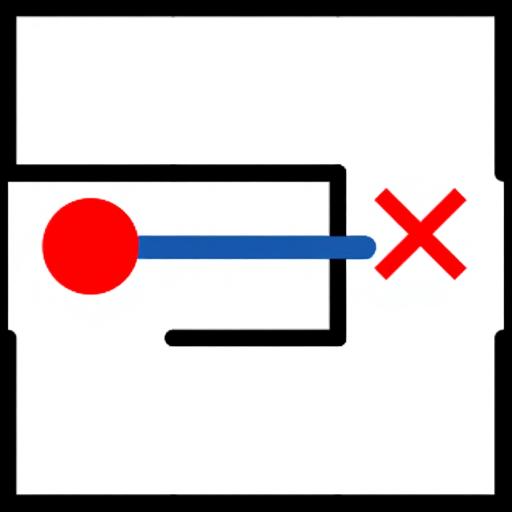}
        \end{subfigure}\hfill
        \begin{subfigure}[b]{0.32\linewidth}
            \centering
            \includegraphics[height=\imgheight, width=\linewidth, keepaspectratio]{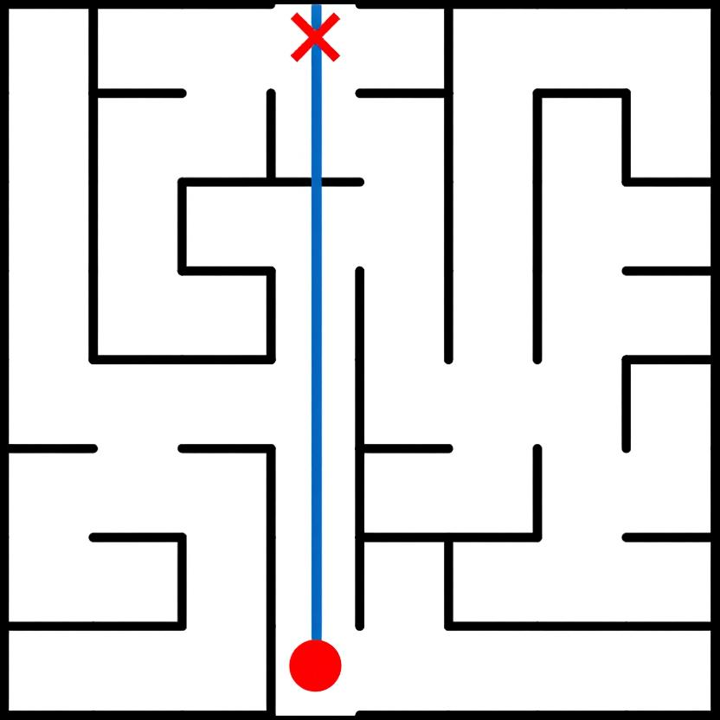}
        \end{subfigure}\hfill
        \begin{subfigure}[b]{0.32\linewidth}
            \centering
            \includegraphics[height=\imgheight, width=\linewidth, keepaspectratio]{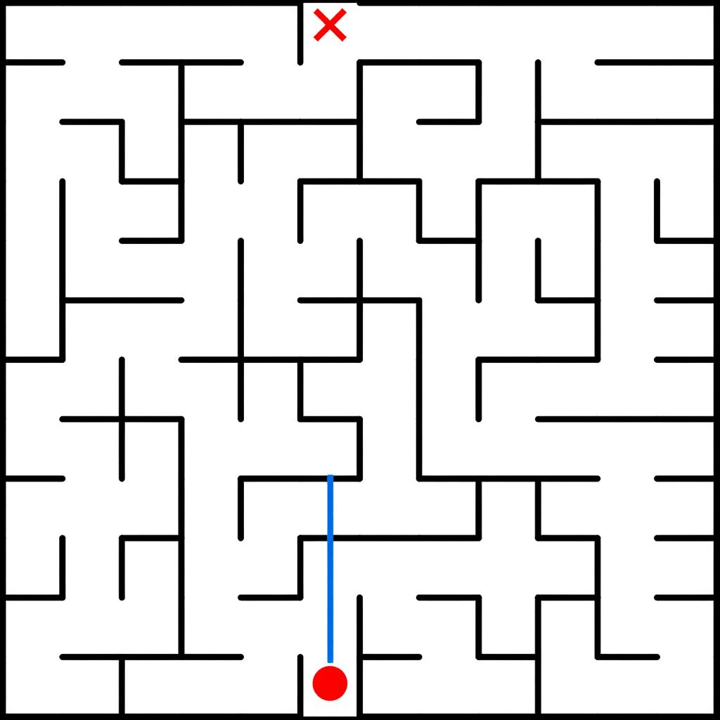}
        \end{subfigure}

        \vspace{0.5em}

        \begin{subfigure}[b]{0.32\linewidth}
            \centering
            \includegraphics[height=\imgheight, width=\linewidth, keepaspectratio]{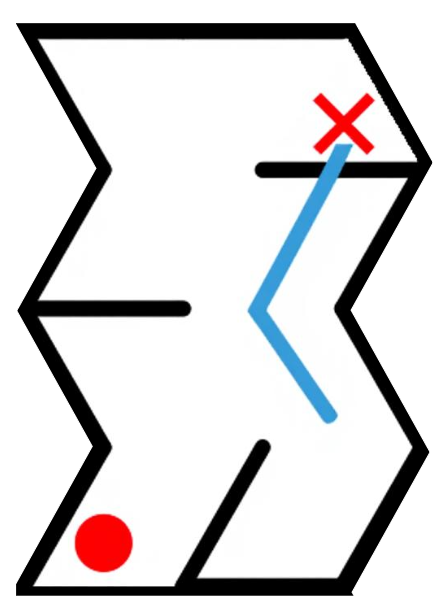}
        \end{subfigure}\hfill
        \begin{subfigure}[b]{0.32\linewidth}
            \centering
            \includegraphics[height=\imgheight, width=\linewidth, keepaspectratio]{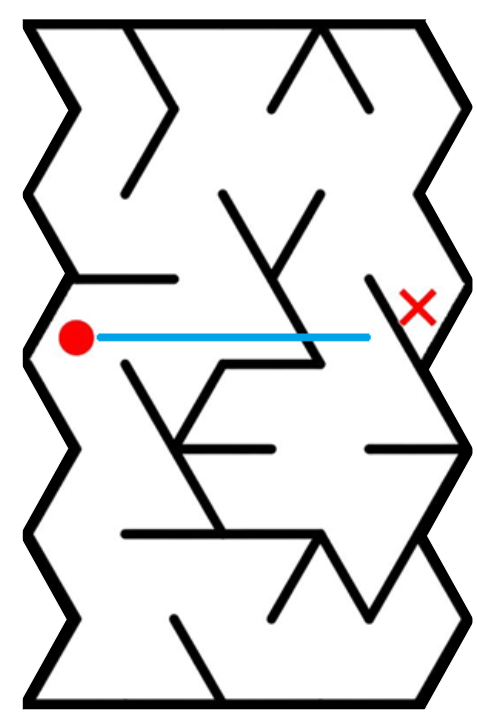}
        \end{subfigure}\hfill
        \begin{subfigure}[b]{0.32\linewidth}
            \centering
            \includegraphics[height=\imgheight, width=\linewidth, keepaspectratio]{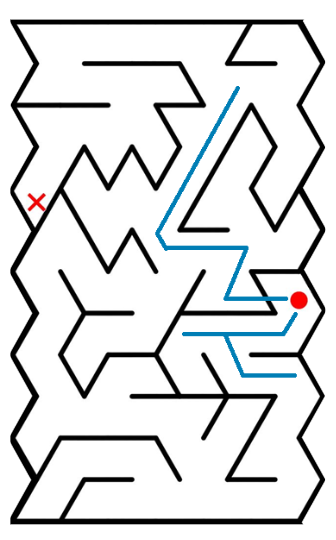}
        \end{subfigure}

        \vspace{0.5em}

        \begin{subfigure}[b]{0.32\linewidth}
            \centering
            \includegraphics[height=\imgheight, width=\linewidth, keepaspectratio]{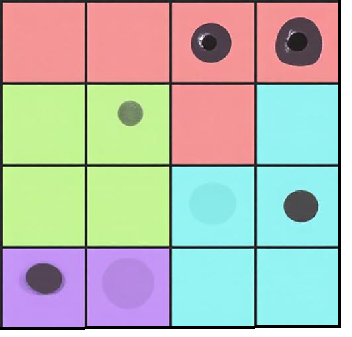}
        \end{subfigure}\hfill
        \begin{subfigure}[b]{0.32\linewidth}
            \centering
            \includegraphics[height=\imgheight, width=\linewidth, keepaspectratio]{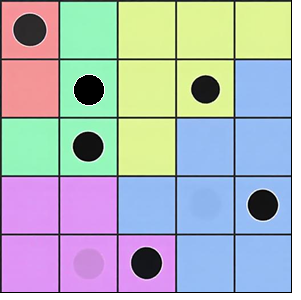}
        \end{subfigure}\hfill
        \begin{subfigure}[b]{0.32\linewidth}
            \centering
            \includegraphics[height=\imgheight, width=\linewidth, keepaspectratio]{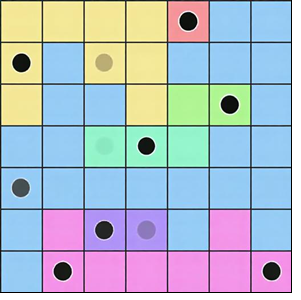}
        \end{subfigure}
        
    \end{minipage}
    \hfill
    \vrule\hfill 
    \begin{minipage}[t]{0.48\textwidth}
        \centering
        \begin{subfigure}[b]{0.32\linewidth}
            \centering
            \includegraphics[height=\imgheight, width=\linewidth, keepaspectratio]{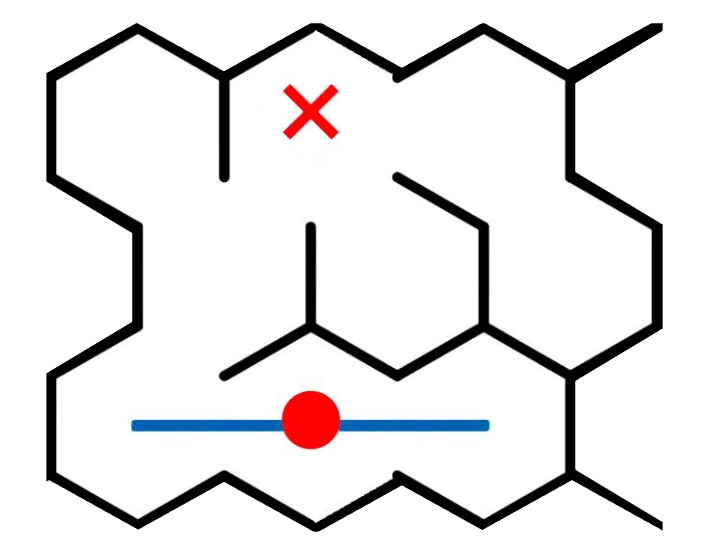}
        \end{subfigure}\hfill
        \begin{subfigure}[b]{0.32\linewidth}
            \centering
            \includegraphics[height=\imgheight, width=\linewidth, keepaspectratio]{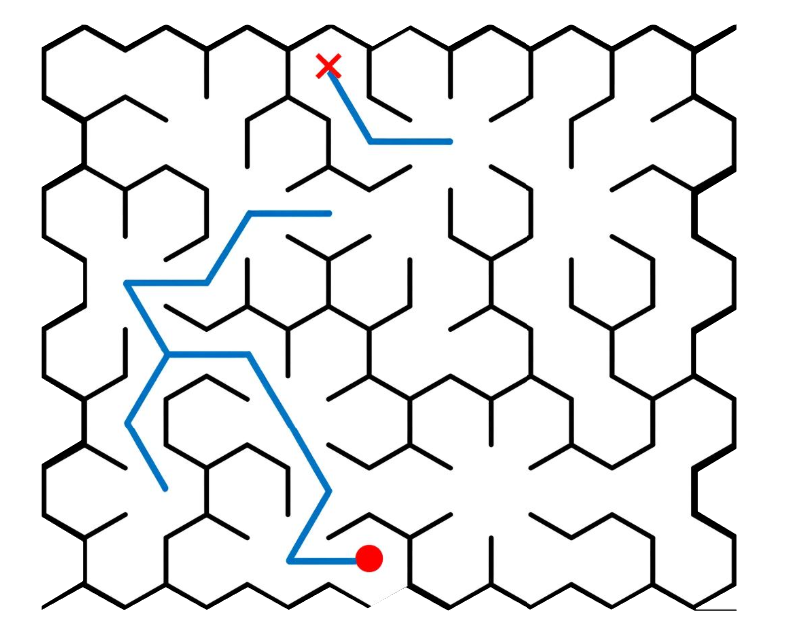}
        \end{subfigure}\hfill
        \begin{subfigure}[b]{0.32\linewidth}
            \centering
            \includegraphics[height=\imgheight, width=\linewidth, keepaspectratio]{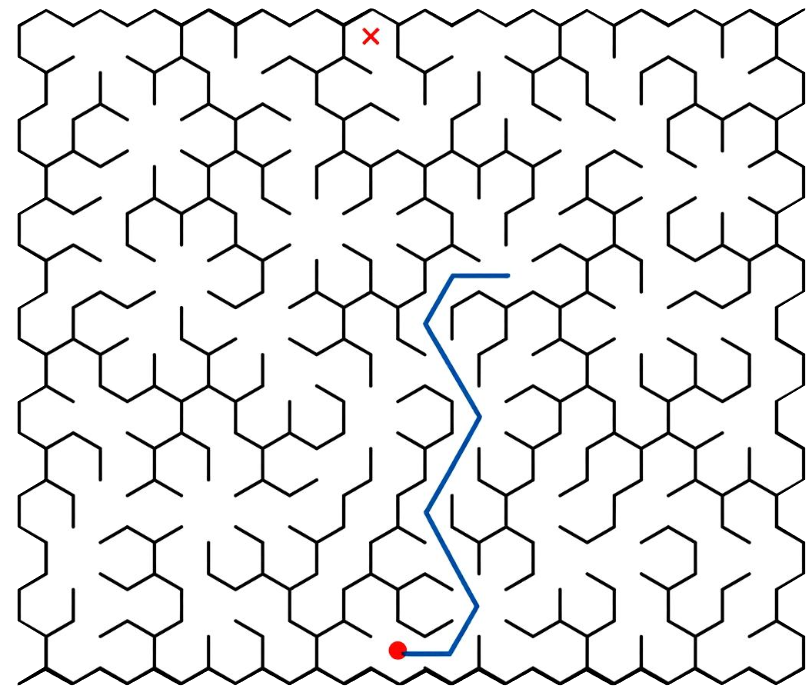}
        \end{subfigure}

        \vspace{0.5em}

        \begin{subfigure}[b]{0.32\linewidth}
            \centering
            \includegraphics[height=\imgheight, width=\linewidth, keepaspectratio]{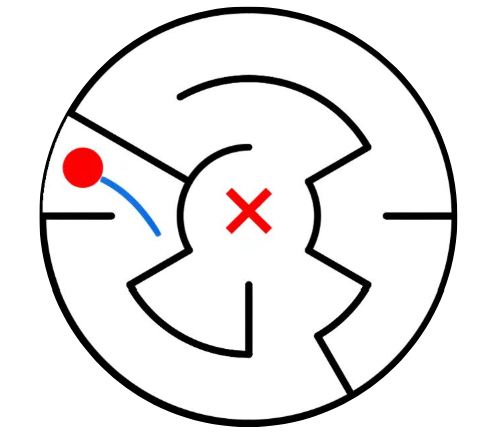}
        \end{subfigure}\hfill
        \begin{subfigure}[b]{0.32\linewidth}
            \centering
            \includegraphics[height=\imgheight, width=\linewidth, keepaspectratio]{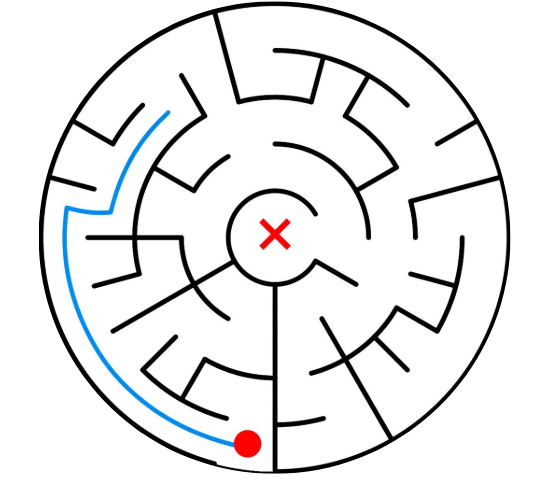}
        \end{subfigure}\hfill
        \begin{subfigure}[b]{0.32\linewidth}
            \centering
            \includegraphics[height=\imgheight, width=\linewidth, keepaspectratio]{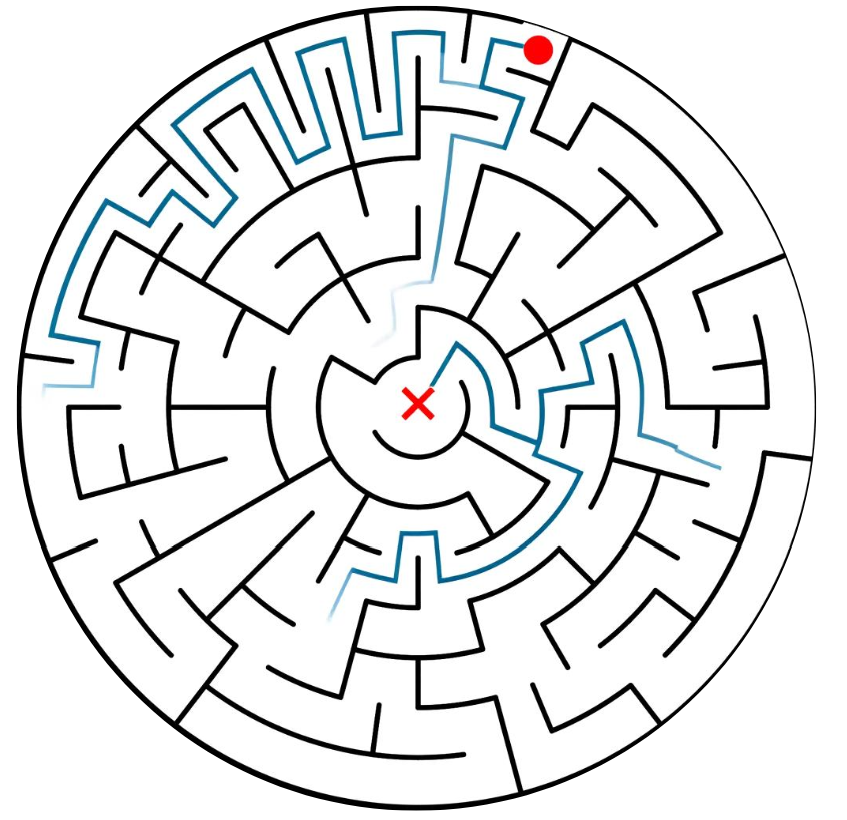}
        \end{subfigure}

        \vspace{0.5em}

        \begin{subfigure}[b]{0.32\linewidth}
            \centering
            \includegraphics[height=\imgheight, width=\linewidth, keepaspectratio]{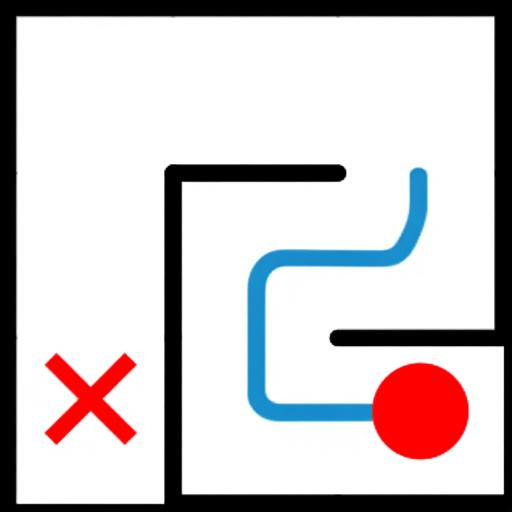}
        \end{subfigure}\hfill
        \begin{subfigure}[b]{0.32\linewidth}
            \centering
            \includegraphics[height=\imgheight, width=\linewidth, keepaspectratio]{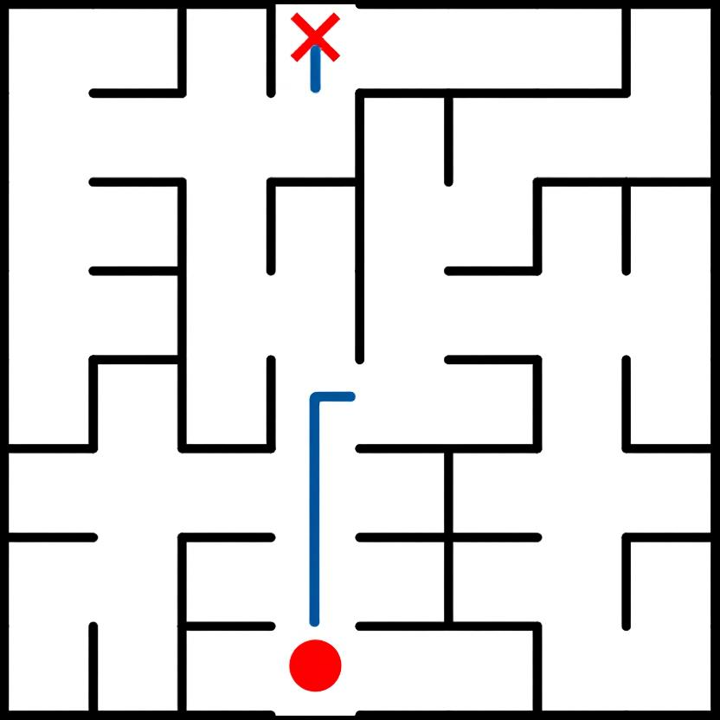}
        \end{subfigure}\hfill
        \begin{subfigure}[b]{0.32\linewidth}
            \centering
            \includegraphics[height=\imgheight, width=\linewidth, keepaspectratio]{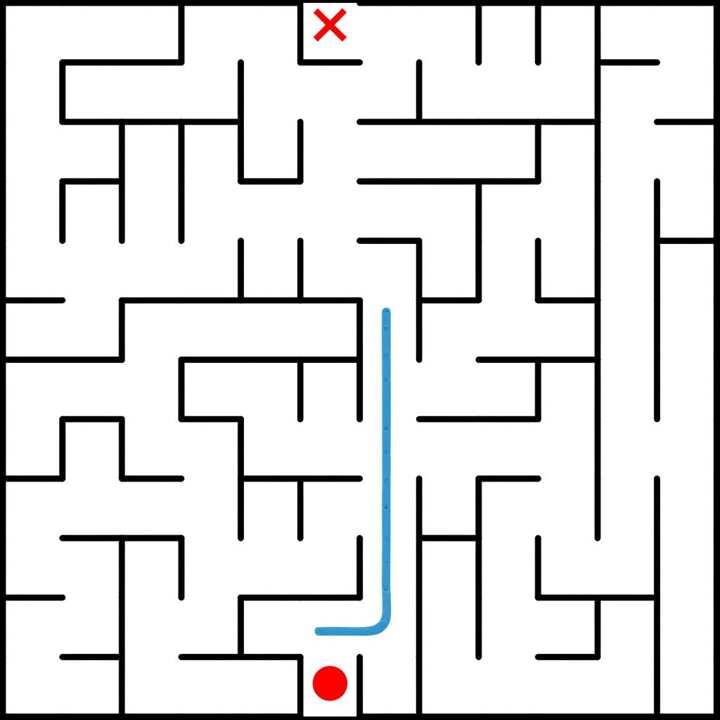}
        \end{subfigure}

        \vspace{0.5em}

        \begin{subfigure}[b]{0.32\linewidth}
            \centering
            \includegraphics[height=\imgheight, width=\linewidth, keepaspectratio]{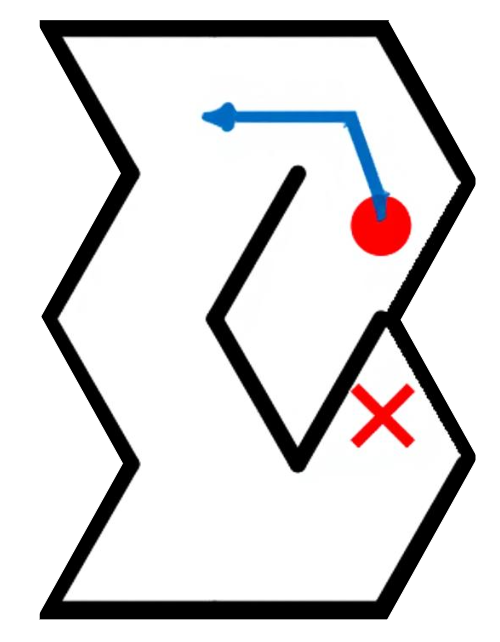}
        \end{subfigure}\hfill
        \begin{subfigure}[b]{0.32\linewidth}
            \centering
            \includegraphics[height=\imgheight, width=\linewidth, keepaspectratio]{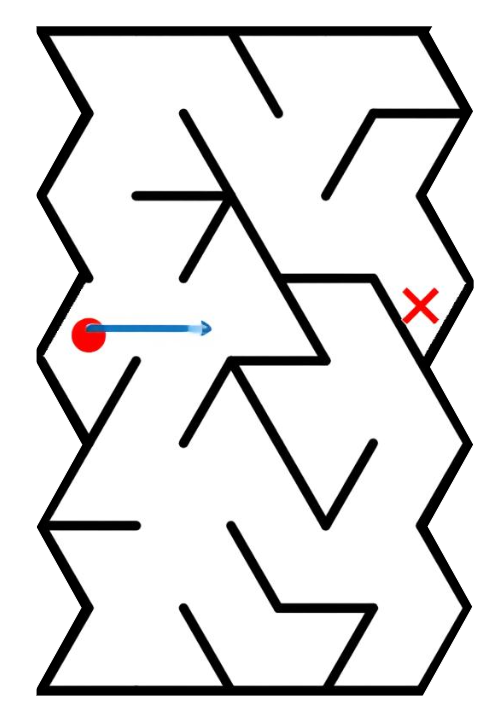}
        \end{subfigure}\hfill
        \begin{subfigure}[b]{0.32\linewidth}
            \centering
            \includegraphics[height=\imgheight, width=\linewidth, keepaspectratio]{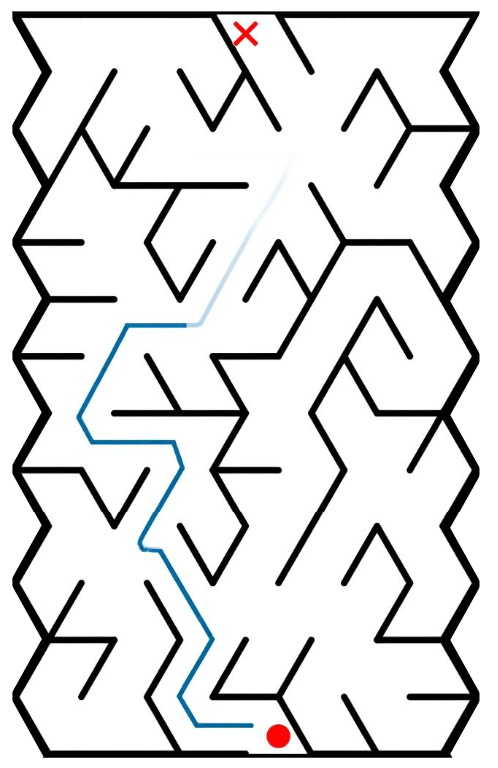}
        \end{subfigure}

        \vspace{0.5em}

        \begin{subfigure}[b]{0.32\linewidth}
            \centering
            \includegraphics[height=\imgheight, width=\linewidth, keepaspectratio]{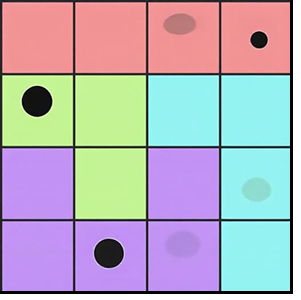}
        \end{subfigure}\hfill
        \begin{subfigure}[b]{0.32\linewidth}
            \centering
            \includegraphics[height=\imgheight, width=\linewidth, keepaspectratio]{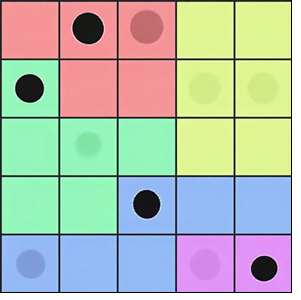}
        \end{subfigure}\hfill
        \begin{subfigure}[b]{0.32\linewidth}
            \centering
            \includegraphics[height=\imgheight, width=\linewidth, keepaspectratio]{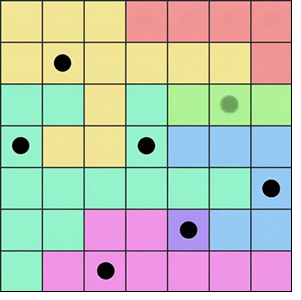}
        \end{subfigure}
    \end{minipage}

    \caption{Fatal cases for $\ishexagon$, $\iscircle$, $\issquare$ and $\istriangle$ mazes, and Queens. \textbf{Left:} boundary violation; \textbf{Right:} incomplete paths.}
    \label{fig:fatal_case_appendix}
\end{figure*}

We further analyze model failures, which can be broadly categorized into two modes: constraint violation and incomplete solution (see Figure~\ref{fig:fatal_case_appendix}). 

\paragraph{Constraint violation} refers to instances where the generated solution fails to adhere to task-specific requirements, reflecting the model's deficit in instruction-following.
On Maze, these violations manifest as invalid trajectories that cross boundaries or connect the start and end points directly--a failure mode that becomes particularly pronounced in complex geometries like $\iscircle$ and $\ishexagon$.
On Queen, this is characterized by erroneous placements that break the global constraint. 

\paragraph{Incomplete solution} refers to cases where the model produces only a partial solution, reflecting a conservative generation strategy.
On Maze, we observe that the model often generates a valid prefix path from the start point but stops early before reaching the end point--a tendency that is particularly pronounced in larger scales or out-of-domain geometries. 
On Queen, this corresponds to instances where the model completes only a subset of goal placements. 
On both tasks, this failure mode results in locally valid but globally incomplete solutions.

\section{Compute Resources and Training Details}

\paragraph{Hardware.}
All open-source experiments were conducted on machines equipped with NVIDIA GPUs. Specifically, we use 4 GPUs of type NVIDIA 5090 with 32 GB memory per GPU for training. For evaluation, we use 2 GPUs of type NVIDIA 5090.

\paragraph{Training Time.}
For main experiment, each model is trained for up to 8 epochs, with early stopping applied based on validation loss. The total training time for each experiment is approximately 24 hours on multi-GPU parallelism.

\paragraph{Inference Cost.}
During evaluation, each sample requires approximately 7.5 seconds for generation in an ideal case.
For the Maze task, the total inference time over the full evaluation set is approximately 5 hours per model.
For the Queen task, the total inference time over the full evaluation set is approximately 4 hours per model.

\paragraph{Implementation Details.}
All open-soure models are implemented using Pytorch 2.9.1. Training is performed with batch size 32, and learning rate 1e-5. 

\section{Comparison with Step-wise Visual Generation}
\label{sec:stepwise}
\label{app:rebuttal_protocols}
We compare EAR with a VPFT-style sequential baseline inspired by Visual Planning~\citep{xu2025letsthink}. Both methods use the same BAGEL initialization, Maze trajectories, prompts, image resolution, training configuration, optimization budget, and evaluation protocol. EAR pairs the initial puzzle with its complete solution; the VPFT-style baseline uses adjacent states from the same Maze trajectories and conditions on its own generated states at inference time. Both are trained on $3\times3$ Hexagon and evaluated on scales $3\times3$--$16\times16$ across all four geometries. Table~\ref{tab:stepwise} separates repeated-evaluation accuracy from calls and latency for a single solution attempt. Hexagon is the training geometry; Circle, Square, and Triangle are unseen geometries. The sequential baseline takes 8.71--17.29 seconds per transition; its measured full-rollout latency ranges from 287.85 to 710.01 seconds per sample.
\begin{table}[t]
\centering
\small
\setlength{\tabcolsep}{5pt}
\ifcsname ver@neurips_2026.sty\endcsname
\captionsetup{position=top,skip=7pt}
\caption{EAR and a matched VPFT-style sequential baseline trained on $3\times3$ Hexagon mazes and evaluated across geometries. Calls and latency describe one solution attempt; sequential call counts and transition latencies are averages. Accuracy uses the repeated-evaluation protocol of the main experiment.}
\label{tab:stepwise}
\fi
\begin{tabular}{llcccccc}
\toprule
& & \multicolumn{2}{c}{Pass@5 (\%) $\uparrow$} & \multicolumn{2}{c}{Calls per solution} & \multicolumn{2}{c}{Seconds per call} \\
\cmidrule(lr){3-4}\cmidrule(lr){5-6}\cmidrule(lr){7-8}
Geometry & Domain & \textbf{EAR} & VPFT-style & \textbf{EAR} & VPFT-style & \textbf{EAR} & VPFT-style \\
\midrule
$\ishexagon$ & ID & 23.29 & 18.71 & 1 & 29.3 & 7.5 & 14.7197 \\
$\iscircle$ & OOD & 8.28 & 0.00 & 1 & 40.7 & 7.5 & 17.2939 \\
$\issquare$ & OOD & 30.00 & 5.71 & 1 & 36.5 & 7.5 & 13.4005 \\
$\istriangle$ & OOD & 40.14 & 5.71 & 1 & 32.7 & 7.5 & 8.7146 \\
\bottomrule
\end{tabular}
\ifcsname ver@neurips_2026.sty\endcsname
\else
\caption{EAR and a matched VPFT-style sequential baseline trained on $3\times3$ Hexagon mazes and evaluated across geometries. Calls and latency describe one solution attempt; sequential call counts and transition latencies are averages. Accuracy uses the repeated-evaluation protocol of the main experiment.}
\label{tab:stepwise}
\fi
\end{table}

Table~\ref{tab:stepwise} shows higher \textsc{Pass@5} for EAR on the training geometry and all three unseen geometries. EAR uses one call (approximately 7.5 seconds), compared with 29.3--40.7 calls and 287.85--710.01 seconds per solution for the baseline, providing a better accuracy--latency trade-off in this setting.

\section{Additional Fine-tuning of Qwen-Image-Edit}
\label{app:qwen_sft}

We additionally adapt Qwen-Image-Edit using LoRA with rank $r=16$ and scaling factor $\alpha=32$, keeping the Maze training data, prompts, test split, and evaluation metrics identical to Table~\ref{tab:main}. As shown in Table~\ref{tab:qwen_sft}, adaptation increases \textsc{Pass@1} from 1.43\% to 35.76\% and \textsc{Pass@5} from 2.14\% to 36.07\%. The gains extend the fine-tuning results to another image-editing model.

\begin{table}[htbp]
\centering
\small
\setlength{\tabcolsep}{5pt}
\ifcsname ver@neurips_2026.sty\endcsname
\captionsetup{position=top,skip=7pt}
\caption{Additional Qwen-Image-Edit results (\%) on AMAZE Maze, using the same data and evaluation protocol as Table~\ref{tab:main}. LoRA uses $r=16$ and $\alpha=32$.}
\label{tab:qwen_sft}
\fi
\begin{tabular}{lrrrrrr}
\toprule
Qwen-Image-Edit & Violation $\downarrow$ & Coverage $\uparrow$ & MSE In $\downarrow$ & MSE Out $\downarrow$ & Pass@1 $\uparrow$ & Pass@5 $\uparrow$ \\
\midrule
Zero-shot & 19.37 & 28.51 & 18.82 & 5.70 & 1.43 & 2.14 \\
\textbf{LoRA fine-tuned} & \textbf{13.00} & \textbf{47.40} & \textbf{10.01} & \textbf{1.38} & \textbf{35.76} & \textbf{36.07} \\
\bottomrule
\end{tabular}
\ifcsname ver@neurips_2026.sty\endcsname
\else
\caption{Additional Qwen-Image-Edit results (\%) on AMAZE Maze, using the same data and evaluation protocol as Table~\ref{tab:main}. LoRA uses $r=16$ and $\alpha=32$.}
\label{tab:qwen_sft}
\fi
\end{table}

\section{Evaluation on the External VFScale Benchmark}
\label{sec:vfscale}
Cross-geometry and cross-scale tests on AMAZE vary the evaluation conditions, but still share a data-generation pipeline. An independently constructed benchmark helps test whether the observed learning and scale extrapolation extend beyond the rendering and sampling choices of our own benchmark. We therefore evaluate on VFScale Maze~\citep{zhang2025vfscale}, whose official data and evaluator provide an external test of the same complete-solution editing formulation.

\paragraph{Setup.}
We fine-tune the original BAGEL checkpoint on the official $4\times4$--$6\times6$ VFScale Maze training sets~\citep{zhang2025vfscale} for 500 steps. We evaluate on its held-out $6\times6$--$15\times15$ test sets using the official evaluator and one generated solution per instance ($N=1$). This experiment uses VFScale training data rather than directly transferring an AMAZE-trained checkpoint.
The reference rows in Table~\ref{tab:vfscale} are taken from the VFScale paper rather than rerun under the BAGEL training configuration.
\begin{table}[htbp]
\centering
\small
\ifcsname ver@neurips_2026.sty\endcsname
\captionsetup{position=top,skip=7pt}
\caption{Success rate (\%) on held-out VFScale mazes with one solution per instance ($N=1$). The first four rows are reference results from Table 1 of \citet{zhang2025vfscale}; EAR-BAGEL is trained on the official $4\times4$--$6\times6$ training sets for 500 steps and evaluated with the official evaluator.}
\label{tab:vfscale}
\fi
\begin{tabular}{lrrrrr}
\toprule
Method & $6\times6$ & $8\times8$ & $10\times10$ & $12\times12$ & $15\times15$ \\
\midrule
Original & 100.00 & 90.62 & 57.81 & 37.50 & 6.25 \\
VFScale training w/o MRNCL & 100.00 & 99.22 & 77.34 & \textbf{56.25} & 25.00 \\
VFScale training w/o KL & 100.00 & 98.44 & 69.53 & 43.75 & 28.12 \\
VFScale training & \textbf{100.00} & \textbf{100.00} & 77.50 & 53.91 & 28.12 \\
\midrule
EAR-BAGEL & 92.59 & 99.11 & \textbf{92.51} & 53.96 & \textbf{36.21} \\
\bottomrule
\end{tabular}
\ifcsname ver@neurips_2026.sty\endcsname
\else
\caption{Success rate (\%) on held-out VFScale mazes with one solution per instance ($N=1$). The first four rows are results from \citet{zhang2025vfscale}; EAR-BAGEL is trained on the $4\times4$--$6\times6$ training sets for 500 steps and evaluated with the official evaluator.}
\label{tab:vfscale}
\fi
\end{table}

\paragraph{Results.}
EAR-BAGEL achieves 92.51\% success on $10\times10$ mazes and 36.21\% on $15\times15$ mazes, exceeding the strongest reported VFScale result at each of these scales by 15.01 and 8.09 percentage points, respectively. At $8\times8$ and $12\times12$, it is within 0.89 and 2.29 points of the best reported results. Performance at $6\times6$ is lower than the reference methods (92.59\% versus 100\%). Since training uses mazes no larger than $6\times6$, the $8\times8$--$15\times15$ results demonstrate scale extrapolation on an independently developed benchmark.
  

\end{document}